\documentclass{article}

\usepackage[T1]{fontenc}

\usepackage{microtype}
\usepackage{graphicx}
\usepackage{subcaption}
\usepackage{booktabs} 

\usepackage{hyperref}

\usepackage[preprint]{icml2026}

\usepackage{amsmath}
\usepackage{amssymb}
\usepackage{mathtools}
\usepackage{amsthm}
\usepackage{colortbl}
\usepackage{tcolorbox}
\usepackage{diagbox}
\usepackage{arydshln}
\usepackage{float}
\usepackage{caption}
\usepackage{adjustbox}
\usepackage{xspace}
\usepackage{marvosym}

\definecolor{lightgray}{gray}{0.85}
\definecolor{lightblue}{rgb}{0.82, 0.95, 0.99}

\usepackage[capitalize,noabbrev]{cleveref}

\theoremstyle{plain}

\theoremstyle{definition}

\theoremstyle{remark}

\usepackage[textsize=tiny]{todonotes}

\icmltitlerunning{Merging Beyond: Streaming LLM Updates via Activation-Guided Rotations}

\begin{document}

\twocolumn[
  \icmltitle{
 Merging Beyond: Streaming LLM Updates via Activation-Guided Rotations
    }



  \icmlsetsymbol{equal}{*}
  \icmlsetsymbol{corr}{}

  \begin{icmlauthorlist}
    \icmlauthor{Yuxuan Yao}{equal,1}
    \icmlauthor{Haonan Sheng}{equal,1}
    \icmlauthor{Qingsong Lv}{2}
    \icmlauthor{Han Wu}{3}{\Letter}
    \icmlauthor{Shuqi Liu}{3}
    \icmlauthor{Zehua Liu}{3}
    \icmlauthor{Zengyan Liu}{1}
    \icmlauthor{Jiahui Gao}{4}
    \icmlauthor{Haochen Tan}{3}
    \icmlauthor{Xiaojin Fu}{3}
    \icmlauthor{Haoli Bai}{3}
    \icmlauthor{Hing Cheung So}{1}
    \icmlauthor{Zhijiang Guo}{5,6}{\Letter}
    \icmlauthor{Linqi Song}{1}{\Letter}
  \end{icmlauthorlist}

  \icmlaffiliation{1}{City University of Hong Kong, Hong Kong SAR}
  \icmlaffiliation{2}{Tsinghua University}
  \icmlaffiliation{3}{Huawei Noah’s Ark Lab, Hong Kong SAR}
  \icmlaffiliation{4}{University of Hong Kong}
  \icmlaffiliation{5}{Hong Kong University of Science and Technology (Guangzhou)}
  \icmlaffiliation{6}{Hong Kong University of Science and Technology}

  \icmlcorrespondingauthor{Han Wu}{wu.han1@huawei.com}
  \icmlcorrespondingauthor{Zhijiang Guo}{zhijiangguo@hkust-gz.edu.cn}
  \icmlcorrespondingauthor{Linqi Song}{linqi.song@cityu.edu.hk}

  \icmlkeywords{Machine Learning, ICML}

  \vskip 0.3in
]



\printAffiliationsAndNotice{}  

\begin{abstract}


The escalating scale of Large Language Models (LLMs) necessitates efficient adaptation techniques. Model merging has gained prominence for its efficiency and controllability. However, existing merging techniques typically serve as post-hoc refinements or focus on mitigating task interference, often failing to capture the dynamic optimization benefits of supervised fine-tuning (SFT). In this work, we propose Streaming Merging, an innovative model updating paradigm that conceptualizes merging as an iterative optimization process.  Central to this paradigm is \textbf{ARM} (\textbf{A}ctivation-guided \textbf{R}otation-aware \textbf{M}erging), a strategy designed to approximate gradient descent dynamics. By treating merging coefficients as learning rates and deriving rotation vectors from activation subspaces, ARM effectively steers parameter updates along data-driven trajectories. Unlike conventional linear interpolation, ARM aligns semantic subspaces to preserve the geometric structure of high-dimensional parameter evolution. Remarkably, ARM requires only early SFT checkpoints and, through iterative merging, surpasses the fully converged SFT model. Experimental results across model scales (1.7B to 14B) and diverse domains (e.g., math, code) demonstrate that ARM can transcend converged checkpoints. Extensive experiments show that ARM provides a scalable and lightweight framework for efficient model adaptation.

\end{abstract}
\section{Introduction}

Guided by scaling laws of Large Language Models (LLM) \cite{DBLP:journals/corr/abs-2505-09388, DBLP:journals/corr/abs-2412-15115, comanici2025gemini25pushingfrontier}, the scale of foundation models continues to expand, driving a growing demand for efficient model development. Model merging, distinguished by its efficiency and controllability, has emerged as a promising research direction for model updating. Recent studies on model merging have focused on data-driven approaches, which differ from conventional model merging in that they leverage calibration data to provide a clearer merging signal. Specifically, \citet{DBLP:journals/corr/abs-2505-14009} found that, with suitable calibration data, a merged model can outperform the best pre-merge model. This naturally raises the question of whether models can be iteratively merged on the calibration set to achieve performance comparable to training on the same data.
Existing continual model merging work \citep{yang2025continual,DBLP:journals/corr/abs-2501-09522} primarily focuses on mitigating task interference during multi-task training. While \citet{DBLP:journals/corr/abs-2505-12082} show that model merging during pretraining can improve training stability and final performance, their approach relies on simple linear interpolation and does not incorporate data-driven adaptation.

In this work, we introduce \textbf{Streaming Merging}, a paradigm that re-envisions model merging as an iterative, geometry-aware optimization process rather than a static interpolation. While our framework applies broadly, we ground our analysis in supervised fine-tuning (SFT), a common setting where practitioners seek to maximize performance under fixed training budgets. Unlike gradient-based training with backpropagation, our \textbf{A}ctivation-guided \textbf{R}otation-aware \textbf{M}erging (ARM) approximates optimization dynamics using only calibration data to steer parameter evolution. We show that conventional arithmetic merging restricts the solution to the affine hull of input checkpoints, imposing a geometric performance ceiling. ARM overcomes this by treating the merging coefficient as an effective learning rate and using activation-derived rotation matrices to approximate gradient-induced directional shifts. This approach enables the merged model to move beyond linear subspaces and navigate the intrinsic geometry of the loss landscape, while also leveraging complementary knowledge latent across input models. Critically, ARM emulates SFT dynamics with only a few initial checkpoints, offering an efficient and practical approach to streaming LLM updates.

\begin{figure*}[t]
    \centering
    \includegraphics[width=0.9\textwidth]{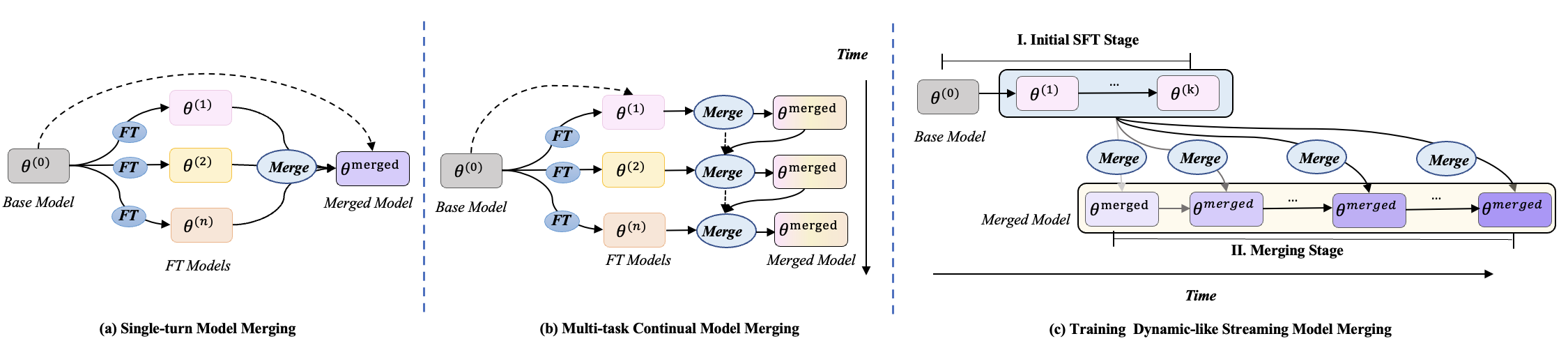}
    \caption{Comparison among different merging pipelines. For Streaming Merging, we select SFT as an illustration.}
    \label{fig:pipeline}
    \vspace{-0.2cm}
\end{figure*} 



Experiments across diverse model scales (1.7B to 14B) and tasks—including mathematical reasoning and code generation—demonstrate that ARM combined with the streaming scheme effectively revitalizes underperforming models from early training stages. Furthermore, it extends performance beyond the convergence of standard training. For instance, in our 7B experiments, ARM achieves \textbf{+3.0} enhancement compared to the final checkpoint; in 14B experiments, ARM yields a \textbf{+1.9} improvement over the fully converged model. We also investigate its applicability in RL scenarios.

We summarize our contributions as: 1) We propose \textbf{Streaming Merging}, a framework that shifts merging from a post-hoc refinement technique to an iterative, training-free optimization process capable of continuously enhancing performance within a consistent task.
2) We present \textbf{ARM}, a non-linear merging operator that addresses the geometric bottlenecks of linear averaging. By aligning semantic subspaces via rotations derived from intermediate activations, ARM preserves the structural integrity of high-dimensional parameter evolution.
3) Empirical evaluations validate the efficacy of ARM within· Streaming Merging across diverse models and tasks, effectively bridging the gap between training-free merging and computationally intensive fine-tuning.

\section{Related Works}

\paragraph{Single-turn Model Merging} Model merging has emerged as an alternative to training-based approaches, enabling the integration of multiple homogeneous models into a unified one \cite{DBLP:journals/corr/abs-2408-07666}. A canonical approach directly averages model parameters \cite{DBLP:conf/siggraph/Shoemake85, DBLP:conf/icml/WortsmanIGRLMNF22}, but this often underperforms due to the non-uniform distribution of features. Task Arithmetic \cite{DBLP:conf/iclr/IlharcoRWSHF23} (TA) addresses this by introducing the task vector, a parameter difference between a pre-trained model and a model fine-tuned for a specific task: \(
\Delta W = W_{\mathrm{sft}} - W_{\mathrm{base}}
\). They demonstrate that simple arithmetic operations on such vectors can effectively edit model behavior. Based on the task-vector, a single, static merged model is formulated as:
\(
W_{\text{merged}} = W_{base} + \frac{1}{N} \sum_{i=1}^{N} \lambda_i \cdot \Delta W_i
\), where $\lambda_i$ is weight coefficient. Subsequent methods like DARE \cite{DBLP:conf/icml/Yu0Y0L24} and TIES \cite{DBLP:conf/nips/YadavTCRB23} refine this paradigm via a pruning-then-scaling process. Additionally, some methods attempt to apply SVD to task vectors \cite{DBLP:journals/corr/abs-2502-10749, lu2024twinmergingdynamicintegrationmodular}, enabling low-rank approximation and noise reduction. POME \cite{DBLP:journals/corr/abs-2510-06627} truncates SVD on the weight difference to balance updates and prune noise. Timber \cite{DBLP:journals/corr/abs-2509-23595} obtains partitioned singular values using eRank as the threshold. AlphaRL \cite{DBLP:journals/corr/abs-2510-00553} predict the final model using low-rank deltas observed from early-training trajectories. 
However, these methods utilize fixed weight coefficient for task vectors, thereby limiting merging efficacy and flexibility. To this end, another line of research, such as AIM \cite{DBLP:journals/corr/abs-2502-02421} and ACM \cite{DBLP:journals/corr/abs-2505-14009}, leverages activations derived from calibration data to determine adaptive coefficients \cite{DBLP:conf/acl/00010HHYS25}. However, these methods rely predominantly on arithmetic merging, constraining the merged model to the affine subspace of task vectors and thereby neglecting the underlying geometric structure. Moreover, despite utilizing calibration data (Fig. \ref{fig:pipeline} (a)), they typically treat merging as a single-step operation rather than a progressive evolution.




\paragraph{Multi-turn Model Merging} 

\citet{DBLP:journals/corr/abs-2505-12082} perform multiple rounds of single-turn merging during pretraining, manipulating checkpoints obtained at different stages of the training process. They employ several methods based on adaptive weight coefficients \cite{DBLP:journals/corr/abs-2209-14981, DBLP:journals/tmlr/Morales-Brotons24, DBLP:conf/uai/IzmailovPGVW18}, all of which are built upon TA. In multi-task learning field, \citet{DBLP:journals/corr/abs-2501-09522} 
proposes continual merging, which advocates sequentially merging checkpoints due to storage constraints or asynchronous availability of task-specific FT models. As shown in Fig. \ref{fig:pipeline} (b), starting from a base model, task-specific models are obtained via sequential fine-tuning; at each step, the model observed in the previous iteration is merged with the model trained on the current new task, yielding a unified, continually evolving multitask model. Existing multi-turn approaches typically treat model merging as a post-hoc step following training or rely on fully fine-tuned models from each task or phase. 
These methods overlook the potential of merging to extend or potentially succeed training. 
Building on this insight, we propose a streaming merging paradigm with principled alignment and scheduling that serves as a viable training-free alternative to conventional optimization.




\section{Methodology}
\subsection{The Streaming Merging Framework} \label{bottleneck}
Distinct from continual merging paradigms designed for multi-task learning, we propose Streaming Merging, a technique tailored for performance enhancement within a consistent task scenario. This approach is applicable across various settings, and is particularly applicable to training-intensive settings, such as supervised fine-tuning and reinforcement learning, where the goal is the continuous refinement towards a target distribution.
Rather than relying on a static final checkpoint or computationally expensive full-model retraining, Streaming Merging leverages a sequence of intermediate candidate checkpoints to refine the model progressively. We formally define the update rule at step $t$ as a function of the previous merged state $W_{prev\_merged}$ and a reference component ${W}_{\text{ref}}$: 
\[
    W_t = \text{StreamingMerge}(W_{prev\_merged}, W_{\text{ref}}),
\]
where ${W}_{\text{ref}}$ is determined by the specific scheduling strategy employed. A natural extension of prior single-turn merging methods to this streaming setting is the \textit{sliding window} recurrence. Assuming a sequence of consecutive checkpoints, this method updates the weights based on the immediate trajectory of the most recent states:
\begin{equation}
W_{t} = W_{t-n} + \lambda/n \sum_{i=t-n}^{t-1}  (W_{i} - W_{t-n}), \quad t > n,
\end{equation}
where $\lambda$ is a scaling coefficient, and $n$ is window size. The initial model $W_{t-n}$ in the sliding window is treated as the reference. While this formulation offers plasticity by rapidly adapting to the latest optimization steps, it relies heavily on transient differences between consecutive checkpoints. Consequently, it is susceptible to trajectory oscillation and error propagation. 
To mitigate this instability, we propose an alternative \textit{anchoring mechanism} that incorporates a global view of the optimization landscape. As illustrated in Fig. \ref{fig:pipeline}(c), we treat a set of $k$ historical checkpoints as fixed anchor references. The update rule is reformulated to pull the current estimate toward the centroid of these anchors: 
\begin{equation}
    W_t = W_{t-1} + \lambda/k \sum_{i=1}^{k} (W_i - W_{t-1}), \quad t > k,
\end{equation}
This formulation interprets the new model as combination of the last merged state and the historical average. We take SFT as an illustration, by grounding the update in the latent $W_i$ optimization direction established during earlier training stages, this approach ensures stability and prevents drift. Ultimately, an ideal framework likely requires a dynamic combination of both strategies. However, whether we use anchors or sliding windows, standard merging methods rely on linear interpolation. This creates a geometric bottleneck. 
For instance, when $\lambda$ is conventionally fixed with $(0,1)$, the anchor merged weights $W_t$ converge to the arithmetic mean of $k$ anchors, $\bar{W}_k = \frac{1}{k} \sum_{i=1}^k W_i$, establishing a hard performance ceiling. Even with adaptive $\lambda_t$, the final model cannot escape the affine hull. Please refer to Appendix \ref{upper bound} for a detailed illustration and theoretical derivation. From an optimization perspective \cite{neyshabur2017geometryoptimizationimplicitregularization, fei2023surveygeometricoptimizationdeep}, gradient descent explores a high-dimensional, curved loss landscape, navigating a nonlinear manifold shaped by architecture, data, and initialization. Therefore, to make the Streaming Merging framework viable, we need a merging operator that can escape this flat subspace.

\begin{figure*}[t]
    \centering
    \includegraphics[width=0.8\textwidth]{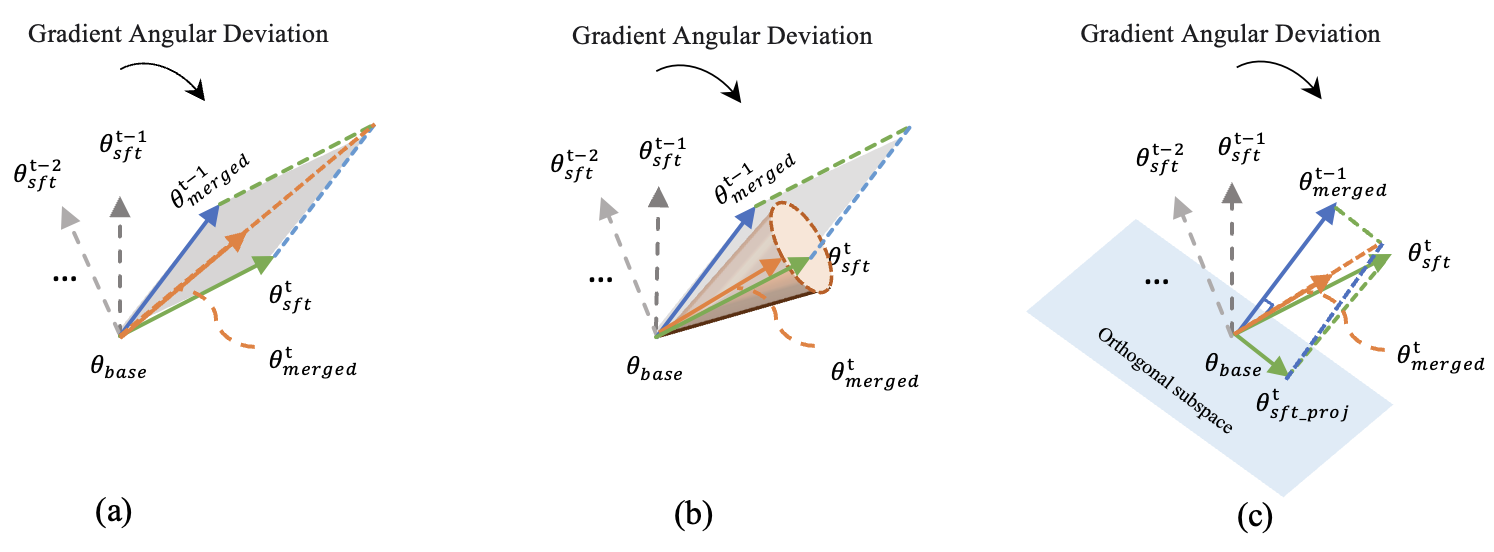}
    \caption{Comparison among Different Merging Approaches. $\theta$ denotes model weights. Gray arrows indicate gradient directions during training, evolving in the full parameter space rather than a planar subspace. (a) TA: The merging outcome is arithmetic. (b) Our ARM: we operate in the full parameter space by rotational alignment, rather than limiting merging to a flat manifold, which is more flexible. (c) OPCM for continual merging: Orthogonal projection is not suitable for training dynamics merging, as checkpoints exhibit complementary rather than mutually exclusive relationships. }
    \label{fig:alg}
\end{figure*}
\vspace{-0.8em}

\subsection{{Simulating Gradient Descent via Activation-guided Rotation: From SFT Perspective}} \label{U_theory}
To address the representational and optimization limitations of arithmetic merging, we advocate for rotation-aware merging. Existing angle-aware schemes such as Slerp merging \cite{DBLP:conf/eccv/JangHSCL24} offer a spherical viewpoint but impose unit-norm normalization on parameters, thereby discarding informative magnitude cues; they are also intrinsically defined only between two endpoints, which complicates extension to three or more checkpoints.
Inspired by the view that activations provide on-the-fly probes of internal features in LLMs \cite{DBLP:journals/corr/abs-2502-02421,DBLP:journals/corr/abs-2505-14009,DBLP:conf/nips/MengBAB22}, and that inter-checkpoint activation differences trace directional shifts in representation space \cite{DBLP:conf/emnlp/PhamN24}, we seek a reliable signal for feature-angle alignment during merging. 

As model merging requires at least two candidate models, we consider checkpoints from early-stage SFT as an illustrative case. This setting is common in practice, where the goal is to effectively optimize a model under limited computational budgets, such that only small-scale training can be performed. We also discuss other initialization scenarios in Appendix \ref{apx:init}.
Let $A=WX$ denote hidden activations for input representations $X$. After SFT, define
\begin{align*}
    \Delta W &:= W_{\mathrm{sft}} - W_{\mathrm{base}} \\
    \Delta A &:= (W_{\mathrm{sft}}X - W_{\mathrm{base}}X) = \Delta WX
\end{align*}
where $A\in\mathbb{R}^{d\times N}$ collects per-sample average hidden activations ($d$ hidden size, $N$ batch size). We compute the full singular value decomposition (SVD) of the activation shift, $\Delta A=U\Sigma V^\top$. The left singular vectors $U\in\mathbb{R}^{d\times d}$ furnish an orthonormal basis over the hidden feature space, capturing the principal modes of activation variation induced by $\Delta W$. Because these directions live in the same output dimension as the weight matrices, they provide a natural target for rotation-based alignment; the right singular vectors $V\in\mathbb{R}^{N\times N}$ span the sample space and are not dimensionally compatible with parameter operations.

We theoretically analyze the relationship between \( U \) and the gradient. Considering the cross-entropy loss $L$ in the SFT stage, the gradient with respect to $W$ is given by:
\begin{align*}
    &\nabla_{W} L = EX^T,\\
    \text{where}~~&E = \frac{\partial L}{\partial A} \in R^{d \times N}.
\end{align*}

Accordingly, for each optimization step (e.g., stochastic gradient descent), the parameter update can be written as
\begin{align*}
    &W_{\mathrm{SGD}}^{t} = W_{\mathrm{SGD}}^{t-1} - \eta \nabla_{W} L, \\
    &\Delta W_{\mathrm{SGD}} = -\eta EX^T
\end{align*}
where $\eta$ denotes the learning rate, which controls the magnitude of each update, while the descent direction is determined by $EX^\top$. When early-stage training checkpoints are used as initial models (e.g., $t=2$, where $t$ denotes the number of optimization steps), we can approximate
\begin{align*}
    &\Delta W_{t} \approx - \eta E_{t}X_{t}^T, \\
    &\Delta A_{t} = \Delta W_{t}X_{t} \approx - \eta E_{t}(X_{t}^TX_{t})
\end{align*}

\noindent Assuming that the input data are independently and identically distributed, we have $X^\top X \approx cI$ for some constant $c$. Consequently,
\begin{align*}
    &\Delta A \approx -c\sum_{t} \eta E_t, \\
    \text{while}~~&\Delta A = U \Sigma V^T.
\end{align*}

Therefore, the left singular vectors $U$ of $\Delta A$ can be interpreted as an empirical estimate of the principal error directions encountered during training.

\subsection{Activation-guided Rotation-aware Model Merging} 
Motivated by the analysis in Section~\ref{U_theory}, we introduce \textbf{A}ctivation-guided \textbf{R}otation-aware \textbf{M}erging (ARM), a data-informed alternative to purely arithmetic merging that preserves magnitude information while aligning updates with the geometry revealed by activations (Fig. \ref{fig:alg} (b)). In its basic form, ARM performs a single-round update as:
\begin{equation}
    W_{\mathrm{merge}}
    \;=\;
    W_{\mathrm{base}}
    \;+\;
    \lambda\, U_{(\mathrm{sft},\mathrm{base})}\!\left(W_{\mathrm{sft}}-W_{\mathrm{base}}\right),
\end{equation}
$W_{\mathrm{sft}}$ denotes a model fine-tuned on the downstream task, and $W_{\mathrm{sft}} - W_{\mathrm{base}}$ constitutes the historical task vector. The orthogonal matrix $U_{(\mathrm{sft},\mathrm{base})}$, derived from activation differences, rotates the task vector into the dominant feature subspace. Compared to TA (Fig. \ref{fig:alg} (a)), our ARM enables merging to transcend the plane (Fig. \ref{fig:alg} (b)). As a norm-preserving transformation, $U$ only adjusts the direction, aligning the update with semantically salient axes. The scalar $\lambda$ determines the update magnitude, functioning analogously to the learning rate in the training process by balancing exploration and stability. Critically, the rotation induced by $U$ provides a gradient-like orientation, steering the task vector toward directions that best capture the principal data-driven variations. Unlike Slerp, which is inherently pairwise and constrained to unit-norm interpolation, ARM naturally extends to multiple checkpoints and mitigates the limitations of arithmetic merging.

We further embed ARM within a dynamic streaming merging framework that combines the stability of anchored initialization with the adaptability of sliding-window refinement. The procedure begins by anchoring to an early SFT checkpoint $W_{\mathrm{i}}$ to faithfully replicate initial optimization dynamics. For the beginning several iterations, updates follow:
\begin{equation}
    W_t = W_{t-1} + \frac{\lambda}{k} \sum_{i=1}^{k} U_{(i,\mathrm{t-1})}\!\left(W_i - W_{\mathrm{t-1}}\right),
\end{equation}
where $\{W_i\}_{i=1}^k$ are finetuned anchor checkpoints and $U_{(i,\mathrm{t-1})}$ aligns each task vector with the activation subspace between $W_i$ and $W_{\mathrm{t-1}}$. Merging successively, we observe diminishing returns due to over-reliance on the initial subspace—a form of representational rigidity that impedes adaptation to later-stage data patterns. To mitigate this, we transition to a sliding-window regime that shifts the reference to the most recent merged model. Empirically, the update becomes:
\begin{equation}
W_{t} = W_{t-n} + \lambda/n \sum_{i=t-n}^{t-1} U_{(i,t-n)} (W_{i} - W_{t-n}), 
\end{equation}
where $U_{(i,t-n)}$ is decomposed from activations of $W_{i}$ and $W_{t-n}$. 
This design mitigates sensitivity to initialization, facilitating seamless adaptation to shifting input distributions while preventing fluctuations and premature convergence. In practice, we partition the calibration data into mini-batches, with each driving one merging iteration. Activations are collected from the recently merged model and relevant references, and rotation matrices are obtained via SVD of their difference. The process terminates when either the angular cosine similarity between consecutive models becomes very close or a maximum iteration budget is reached. This two-stage schedule preserves geometric fidelity to early training while retaining flexibility for late-stage refinement, enabling ARM to go beyond conventional merging and emulate fine-tuning dynamics.

\section{Experiments}
\begin{table*}[ht!]
\fontsize{9}{10} \selectfont
\centering
\caption{Evaluations of different methods on Qwen models. The superior results of the merging methods on each benchmark are highlighted in bold. (SW) refers to vanilla sliding window mechanism mentioned in Section 3.1. Since inputs were wrapped with a chat template during fine-tuning, all evaluations set \textit{$apply\_chat\_template=True$}. For Qwen3-1.7B-Base, we additionally evaluated the setting without an instruct template, yielding scores of 77.3, 41.9, 10.5, 27.0, and 17.6 on the respective benchmarks, with an average performance of 34.9.}
\label{tab:qwen7b_eval}
\begin{tabular}{lcccccc}
\toprule[0.8pt]
\diagbox{Method}{Bench} & GSM8K & MATH500  & Olympiad Bench & College Math  & Minerva Math & Avg. \\
\hline
\rowcolor{lightblue}
\multicolumn{7}{c}{\textit{Qwen3-1.7B-Base}} \\
Qwen3-1.7B-Base  &23.0 &15.0 &6.1 &7.6 &4.4 &11.2 \\ \hline
\rowcolor{lightgray}
\multicolumn{7}{c}{\textit{Training-based Methods}} \\
Checkpoint10 &58.4 &44.0 &18.5 &24.5 &10.3 &31.1 \\ 
Checkpoint20 &65.0 &51.4 &15.9 &33.9 &19.1 &37.1 \\ \cdashline{1-7}
Checkpoint150 &76.0	&55.2 &19.4	&\textbf{38.1} &20.2	&41.8\\
LoRA  &75.0	&54.4 &\textbf{24.4}	&35.6 &15.4 &40.9 \\ \hline
\rowcolor{lightgray}
\multicolumn{7}{c}{\textit{Streaming Merging-based Methods}} \\
TA &71.2 &53.0 &23.6 &33.3 &15.8 &39.4 \\
TIES &66.6 &53.2 &22.1 &29.8 &12.1 &36.8 \\
DARE &67.5 &53.8 &21.5 &35.9 &20.2 &39.8 \\ \cdashline{1-7}
AIM &59.2 &46.8 &16.3 &30.9 &18.4 &34.3 \\
ACM &69.9 &\textbf{57.2} &22.7 &35.9 &18.0 &40.7 \\
\hline
ARM  &\textbf{76.2} &55.2 &23.6 &33.8 &\textbf{23.9} &\textbf{42.5}  \\
\hline
\rowcolor{lightblue}
\multicolumn{7}{c}{\textit{Qwen2.5-7B}} \\
Qwen2.5-7B  &80.9 &58.6 &25.4 &38.6 &18.3 &44.4 \\ \hline
\rowcolor{lightgray}
\multicolumn{7}{c}{\textit{Training-based Methods}} \\
Checkpoint10 &82.2 &66.3 &29.9 &41.3 &26.8 &49.3 \\ 
Checkpoint20 &84.9 &64.4 &27.1 &44.7 &29.4 &50.1 \\ \cdashline{1-7}
Checkpoint100 &86.3 &65.3 &27.0 &44.0 &33.2 &51.2 \\
LoRA &\textbf{88.9} &67.0 &31.0 &44.9 &29.4 &52.2 \\ \hline
\rowcolor{lightgray}
\multicolumn{7}{c}{\textit{Streaming Merging-based Methods}} \\
TA &86.1 &69.4 &\textbf{32.9} &45.2 &35.3 &53.7 \\
TIES &87.8 &69.2 &32.7 &\textbf{46.0} &31.6 &53.5 \\
DARE &87.4 &67.0 &29.2 &44.2 &34.6 &52.5 \\ \cdashline{1-7}
AIM &85.4 &64.6 &28.9 &43.3 &24.6 &49.4 \\
ACM &87.0 &68.2 &30.5 &45.1 &\textbf{36.4} &53.4 \\ \hline
ARM &87.9 &\textbf{69.6} &31.8 &\textbf{46.0} &35.7 &\textbf{54.2} \\\hline
\rowcolor{lightblue}
\multicolumn{7}{c}{\textit{Qwen2.5-14B}} \\
Qwen2.5-14B  &89.4 &61.4 &28.3 &40.4 &25.0 &48.9 \\ \hline
\rowcolor{lightgray}
\multicolumn{7}{c}{\textit{Training-based Methods}} \\
Checkpoint10 &91.7 &71.2 &35.4 &46.1 &\textbf{43.0} &57.5 \\ 
Checkpoint20 &92.6 &71.4 &36.0 &45.9 &42.3 &57.6 \\ \cdashline{1-7}
Checkpoint100 &92.3 &71.4 &35.9 &46.1 &41.2 &57.4 \\
\rowcolor{lightgray}
\multicolumn{7}{c}{\textit{Streaming Merging-based Methods}} \\
TA &\textbf{92.9} &73.8 &\textbf{40.0} &46.7 &42.3 &59.0 \\
ARM  &92.4 &\textbf{75.2} &39.3 &\textbf{47.2} &42.3 &\textbf{59.3}  \\
\bottomrule[0.8pt]
\end{tabular}
\end{table*}
We focus our primary evaluation on SFT. Extended experiments on RL are presented in Section~\ref{Further Analyses} and Appendix~\ref{7b_rl}, while LoRA-based results are provided in Appendix~\ref{qwen2.5_7b_lora}.
\paragraph{Models}
We conducted experiments across various model scales and architectures, including Qwen2.5-7B, Qwen2.5-14B \cite{DBLP:journals/corr/abs-2412-15115}, Qwen3-1.7B-Base \cite{DBLP:journals/corr/abs-2505-09388}, and LLaMA3.2-3B-Instruct \footnote{https://huggingface.co/meta-llama/Llama-3.2-3B-Instruct}.  To ensure reproducibility, we employed the public evaluation toolkit provided by QwenLM \footnote{https://github.com/QwenLM/Qwen2.5-Math} for the Qwen series, adhering to their recommended versions of dependencies. We set the sampling temperature to 0.7, top-$p$ to 0.8, and the maximum generation tokens to 4096. All reported values represent the average of 5 independent evaluations.

\paragraph{Datasets and Benchmarks}
We utilize \textsc{NuminaMath}~\cite{numina_math_datasets} as training and calibration data, a synthetic mathematical reasoning dataset of variable reliability. We regenerate its reasoning traces using \textsc{Gemini}-2.5-Pro~\cite{comanici2025gemini25pushingfrontier} and validate their correctness with \textsc{GPT}-5\footnote{\url{https://openai.com/gpt-5/}}. Only samples that pass automated consistency and logical validity checks are retained, yielding a cleaned training set of approximately $6.5K$ examples, which we will publicly release. For evaluation, we adopt a comprehensive suite of mathematical benchmarks covering diverse difficulty levels and domains: GSM8K \cite{DBLP:journals/corr/abs-2110-14168}, MATH500 \footnote{https://huggingface.co/datasets/HuggingFaceH4/MATH-500}, OlympiadBench \cite{DBLP:conf/acl/HeLBHTSHHHZLQL024}, CollegeMath \cite{DBLP:conf/icml/TangZWW24}, and Minerva-Math \cite{lewkowycz2022solvingquantitativereasoningproblems}. Additionally, to assess generalization beyond mathematics, we include a code reasoning analysis in Section~\ref{Further Analyses}, demonstrating the generalizability of our merging framework.

\paragraph{Baselines} We evaluate the effectiveness of our ARM method by comparing it with training approaches and prevalent model-merging techniques, including full-parameter training on the entire dataset, LoRA-based training \cite{DBLP:conf/iclr/HuSWALWWC22}, TA \cite{DBLP:conf/iclr/IlharcoRWSHF23}, TIES-Merging \cite{DBLP:conf/nips/YadavTCRB23}, DARE \cite{DBLP:conf/icml/Yu0Y0L24}, data-driven AIM \cite{DBLP:journals/corr/abs-2502-02421}, and ACM \cite{DBLP:journals/corr/abs-2505-14009}. LoRA fine-tunes large models by optimizing low-rank updates to frozen pretrained weights. TA adopts arithmetic operations on task-specific parameter offsets. 
TIES enhances TA by aligning signs, pruning inconsistent dimensions, and interpolating only the consensus subspace. DARE sparsifies task vectors via random masking and restores magnitude through scaling to suppress noise and emphasize robust directions. Leveraging calibration data, AIM emphasizes the activation of the base model to prevent catastrophic forgetting. ACM utilizes activation-based mutual information to observe adaptive weight coefficients. The batch size is set to 64. More hyperparameters can be found at Appendix~\ref{hyperparameters}.

We omit comparison with multi-task continual merging methods like OPCM  \cite{DBLP:journals/corr/abs-2501-09522} , as the fine-tuned checkpoints exhibit substantial semantic overlap. In such cases, merging strategies based on orthogonal projection, which are designed under the assumption of task orthogonality, risk attenuating shared optimization signals and thereby eroding training-induced improvements (cf.~Fig.~\ref{fig:alg}(c)).

\paragraph{Main Results} 

Experiments on the Qwen series are presented in Table \ref{tab:qwen7b_eval}, and results for LLaMA are provided in Appendix~\ref{llama}. We collect a training checkpoint every 10 steps and then perform streaming merging for all merging operators using the first two obtained checkpoints. According to Table \ref{tab:qwen7b_eval}, the following observations can be derived:

\textbf{(1) The Streaming Merging framework is highly versatile and consistently enhances model performance.} Experiments across models of varying scales and architectures show that our streaming pipeline effectively leverages training checkpoints by seamlessly integrating diverse single-turn merging strategies, including TA, ACM, and ARM. For example, on Qwen2.5-7B, the converged checkpoint (checkpoint 100) attains 51.2\% accuracy, whereas most streaming merging variants (TA, TIES, and ARM) exceed \textbf{53.5\%}, with the best result reaching \textbf{54.2\%}. We defer the per-step performance evolution to the Appendix \ref{envolve}, and analyze the impact of using the sliding window strategy and the anchoring strategy separately in the Appendix \ref{abl_sl_achoring}. In addition, under the streaming merging setting, we observe that methods such as TA typically require 2 to 5 iterations, whereas ARM with the rotation mechanism requires 3 to 6 iterations. This finding suggests that structurally informed adjustments in the weight space incur a modest increase in computational cost while enabling finer alignment with task-specific semantic directions, which in turn improves the performance of the merged model.

\textbf{(2) ARM integrated with a streaming scheme consistently achieves the strongest performance across model scales.} ARM reaches 42.5 on Qwen3-1.7B-Base, 54.2 on Qwen2.5-7B, and 59.3 on Qwen2.5-14B, outperforming TA, ACM, and even finetune-based methods without any additional training. This underscores ARM’s effectiveness in iteratively refining early checkpoints through dynamic, streaming-aware merging. As shown in Fig~\ref{fig:alg} (b), compared to arithmetic interpolation, ARM incorporates angular alignment during merging, better mimicking the directional shifts induced by gradient descent during training while effectively integrating complementary knowledge latent across checkpoints, thereby yielding superior merging performance.

\textbf{(3) ARM integrated with streaming scheme can intervene early in training.} On Qwen3-1.7B and Qwen2.5-7B, training has not yet converged at checkpoint 20, continued training often yields diminishing returns and may lead to suboptimal convergence, vanishing gradient signals, or mild overfitting. In contrast, our ARM effectively replaces further SFT by merging only early checkpoints. It employs an activation-guided rotation mechanism to perform structure-aware alignment of parameters from different checkpoints, thereby integrating complementary knowledge acquired across training stages. By aligning task-relevant directions through geometric transformations, ARM produces models that surpass the fully converged SFT checkpoints by \textbf{+0.7} and \textbf{+3.0} accuracy points, respectively. Appendix~\ref{rotat_mag} further illustrates that the parameter trajectories produced by ARM closely emulate those of standard fine-tuning, suggesting its ability to imitate training dynamics geometrically. We further observe that AIM underperforms on both scales, likely due to its conservative constraint—limiting deviation from the base model to mitigate catastrophic forgetting—at the expense of performance gains.


\textbf{(4) ARM integrated with streaming scheme continues to improve performance even after training has fully converged.} On Qwen2.5-14B, where SFT checkpoint plateaus at about 57.5, our method further merges these checkpoints to achieve 59.3 (\textbf{+1.8}). This indicates that, despite convergence, checkpoints retain complementary knowledge; by leveraging rotation-aware merging, ARM better aligns semantic directions in weight subspaces, surpassing the performance ceiling of the converged model. Moreover, we observe that streaming merging typically requires $3\sim6$ iterations when applied prior to convergence, but only $2\sim3$ iterations once performance has plateaued. Besides, the sliding-window annealing strategy is also unnecessary at this stage. This is because early checkpoints exhibit greater divergence in optimization trajectories, necessitating more iterations to align their representations, whereas late-stage checkpoints have converged to a stable region of the loss landscape and therefore require fewer adjustments for effective merging. Additionally, we also evaluate ARM in the single-turn merging setting; please refer to the Appendix~\ref{single ARM} for details.

\section{Further Analyses} \label{Further Analyses}

\begin{figure}[htbp]
    \centering
    \includegraphics[width=\linewidth]{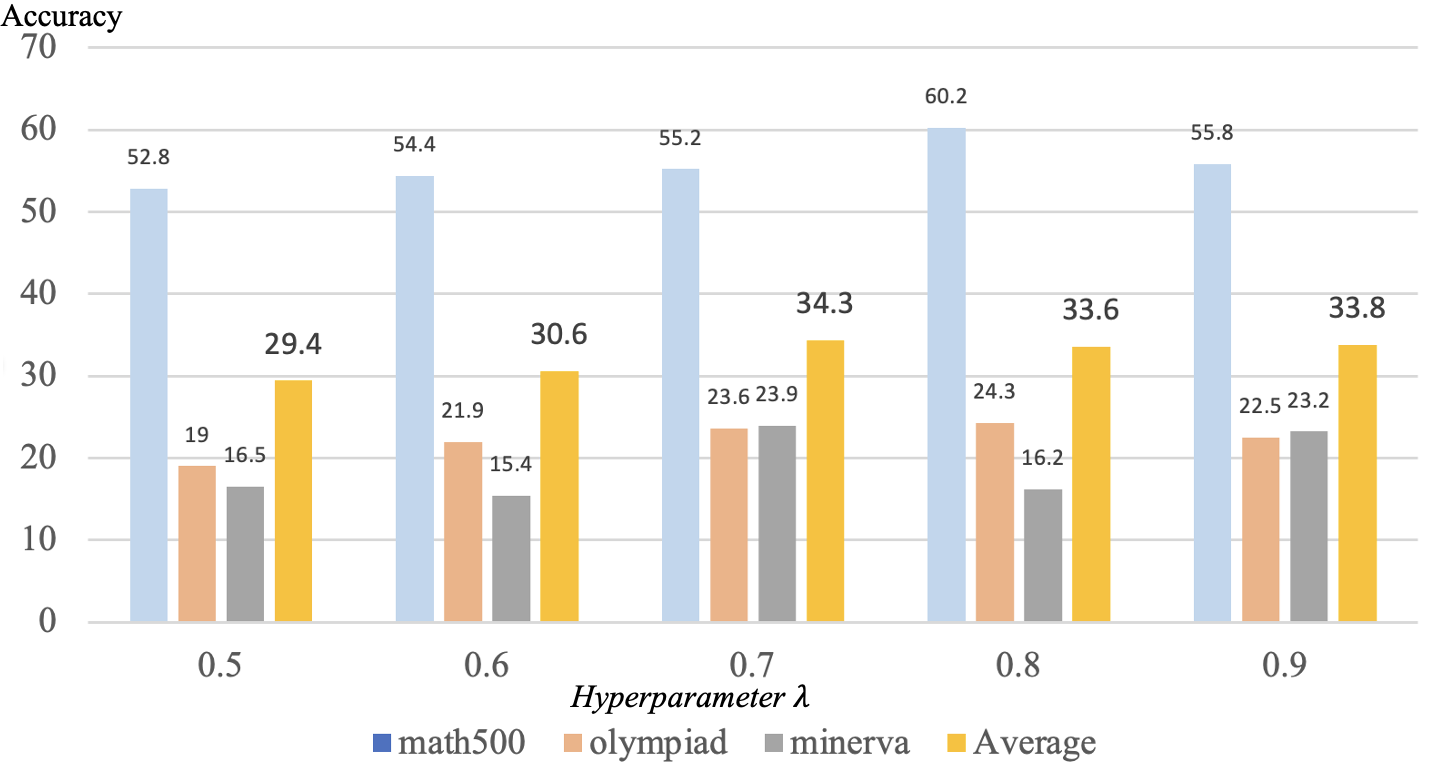}
    \caption{Ablation Study on Hyperparameter $\lambda$}
    \label{fig:abl1}
\end{figure}
\vspace{-0.8em}

\paragraph{Ablation Study on Hyperparameter $\lambda$, Checkpoint Counts, and Batch Size} To investigate the sensitivity of hyperparameter $\lambda$, model counts, and  batch size, we conducted ablation studies on Qwen2.5-7B to analyze their impact. Fig. \ref{fig:abl1} shows that our method is robust to the hyperparameter $\lambda$, with performance stabilizing for $\lambda \geq 0.7$ and remaining insensitive. Table \ref{qwen2.5-7b-counts} results show that merging the base model with two candidate models already achieves performance saturation, and further increasing the number of models yields no significant improvement. Beyond our ARM method, other merging approaches exhibit similar trends. We therefore recommend that streaming merging be performed using the base model with one or two candidate models; adding more candidates tends to introduce redundancy without significant gain.

\begin{figure}[htbp]
    \centering
    \includegraphics[width=0.86\linewidth]{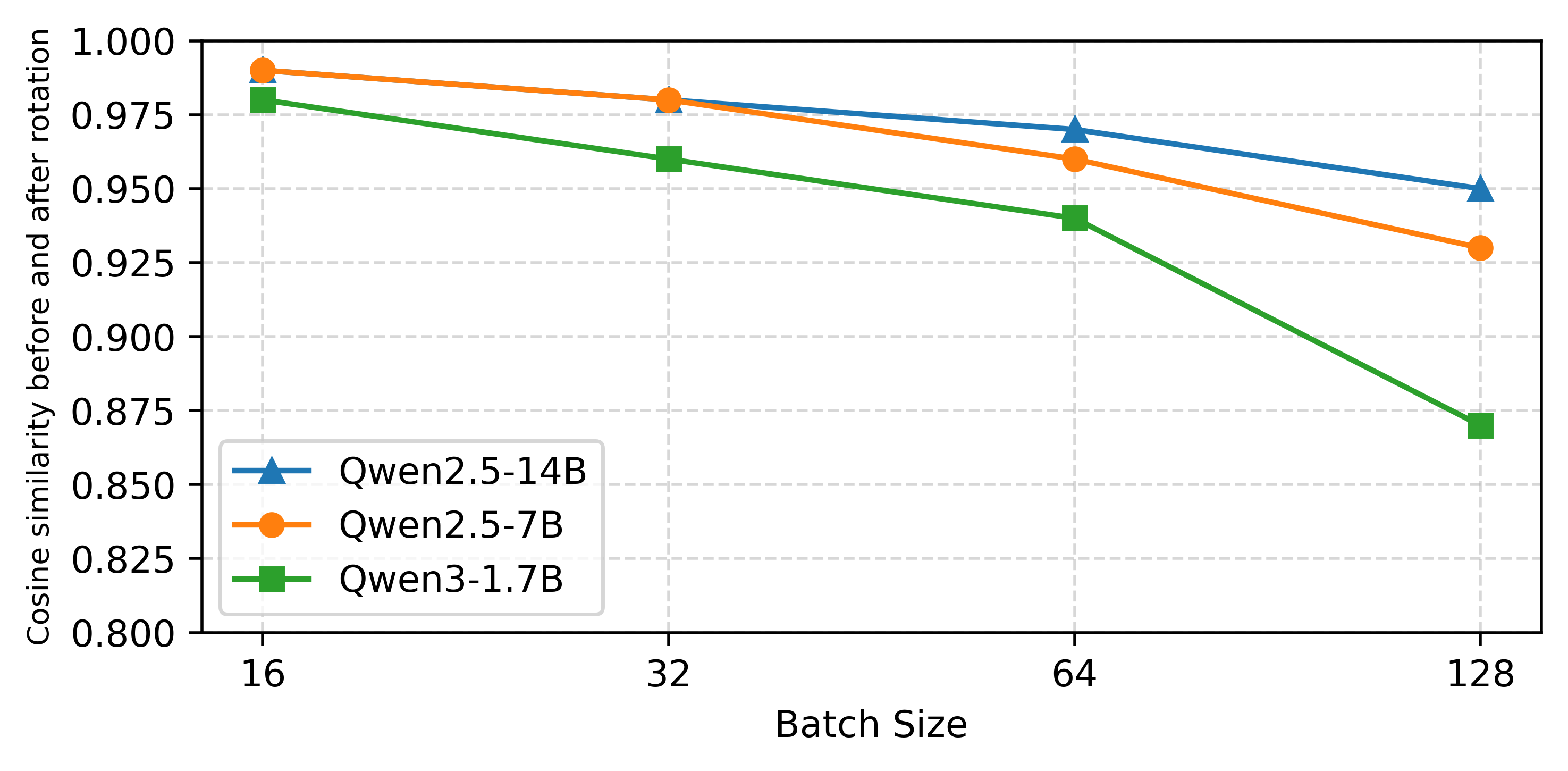}
    \caption{Ablation Study on batch size, we analyze the cosine similarity of the weights before and after rotation. }
    \label{fig:bsz}
\end{figure}
\vspace{-0.8em}

It can be observed in Fig. \ref{fig:bsz} that as the batch size increases, cosine similarity gradually decreases. This aligns with theoretical expectations \cite{vershynin2018high}: due to the independence of data samples, larger batch sizes introduce greater randomness, and high-dimensional random vectors tend to become orthogonal. Furthermore, the larger model exhibits lower sensitivity to batch size due to the larger hidden size. Overall, the maximum reduction in cosine similarity is approximately 0.1, demonstrating the robustness of our method. In practice, we suggest using a batch size of 64. Additionally, ablation studies of sliding window and anchoring scheme are presented at Appendix \ref{abl_sl_achoring}.

\begin{table}
\fontsize{9}{10} \selectfont
\caption{Ablation study on the number of merged checkpoints}
\label{qwen2.5-7b-counts}
\begin{tabular}{ccccc}
\toprule[0.8pt]
Ckpt Counts & MATH500 & Olympiad & Minerva & AVG  \\ \hline
1 &68.2 &30.7 &29.1 &42.7\\
2 &71.6 &33.6 &33.8 &46.3\\
3 &70.8 &33.5 &34.9 &46.4\\

 \bottomrule[0.8pt]
\end{tabular}
\end{table}


\begin{table}
\fontsize{9}{10} \selectfont
\caption{Performance of \textit{Streaming Merging} on Qwen3-1.7B-Base after RL training via GRPO algorithm. ckpt repersents checkpoint .}
\label{qwen3-1.7b-rl}
\begin{tabular}{lcccc}
\toprule[0.8pt]
Qwen3-1.7B & MATH500 & Olympiad & Minerva & AVG  \\ \hline
RL-ckpt 40         & 65.6    & 25.9           & 31.6         & 41.0 \\
RL-ckpt 50         & \textbf{67.1}    & 27.9           & 30.9         & 41.9 \\ \cdashline{1-5}
TA             & 65.6    & \textbf{28.4}           & 30.9         & 41.6 \\
ARM          & 66.3    & \textbf{28.4}           & \textbf{32.4}         & \textbf{42.4} \\ \bottomrule[0.8pt]
\end{tabular}
\end{table}
\vspace{-0.8em}
\paragraph{Extended Experiments via Reinforcement Learning} The success of Streaming Merging in SFT naturally motivates its extension to early-stage RL, where dynamic merging of intermediate checkpoints could emulate the full optimization trajectory and reduce training cost. Our empirical investigation is primarily based on the GRPO algorithm \cite{DBLP:journals/corr/abs-2402-03300}. However, preliminary experiments on Qwen3-1.7B-Base and Qwen2.5-7B do not yield the same performance gains observed in SFT—likely due to the high variance and non-stationarity of RL: noisy policy gradients and sparse rewards hinder stable semantic alignment across checkpoints. Nevertheless, after RL convergence, ARM
integrated with Streaming Merging applied to the final few checkpoints (checkpoint40, checkpoint50) further boosts performance, leveraging their complementary knowledge. As shown in Table \ref{qwen3-1.7b-rl}, 2$\sim$3 merging iterations on the 1.7B model surpass the fully trained RL policy. Similar results for the 7B model are provided in the Appendix \ref{7b_rl}. 

\begin{table}
\fontsize{9}{10} \selectfont
\caption{Performance of ARM combined with RL editing methods on Qwen3-1.7B-Base}
\label{qwen3-1.7b-rl-editing}
\begin{tabular}{lcccc}
\toprule[0.8pt]
Qwen3-1.7B & MATH500 & Olympiad & Minerva & AVG  \\ \hline
Timber &64.8 &28.9 &28.3 &40.6  \\
$+$ ARM &65.8 &29.3 &30.5 &41.9  \\ \cdashline{1-5}
POME   &65.0 &27.3 &28.7 &40.3     \\
$+$ ARM &66.6 &28.7 &27.7 &40.6\\
\bottomrule[0.8pt]
\end{tabular}
\end{table}
Moreover, recent RL editing methods, such as Timber and POME have revealed that improvements in reasoning ability during RL are almost entirely governed by low-rank subspaces of parameter updates \cite{DBLP:journals/corr/abs-2510-06627, DBLP:journals/corr/abs-2509-23595, DBLP:journals/corr/abs-2510-00553}. Our ARM framework seamlessly integrates with prevailing RL editing techniques, enabling low-rank edits to better approximate full parameter performance. As shown in the Table \ref{qwen3-1.7b-rl-editing}, applying ARM’s rotation-based alignment to the weight delta prior to low-rank editing suffices to yield significant performance gains. 

\begin{table}
\fontsize{9}{10} \selectfont
\centering
\caption{Performance of \textit{Streaming Merging} on Qwen2.5-7B across code benchmarkss, where LiveCode indicates LiveCodeBench, HE-Pro represents HumanEval-Pro.}
\label{qwen2.5-7b-code}
\begin{tabular}{lcccc}
\toprule[0.8pt]
Method & LiveCode & HE-Pro & MBPP-Pro & AVG  \\ \hline
\rowcolor{lightgray}
Qwen2.5-7B &2.5 &47.6 &59.2 &36.4 \\
ckpt10  &12.2 &51.8 &59.8 &41.3   \\ 
\rowcolor{lightgray}
ckpt20 &24.6 &56.7 &59.3 &46.9 \\
ckpt100 &24.6 &55.5 &59.3 &46.4 \\ \cdashline{1-5}
TA &25.3 &57.9 &\textbf{60.9} &48.0 \\
ARM &\textbf{25.4} &\textbf{58.1} &60.8 &\textbf{48.2} \\
\bottomrule[0.8pt]
\end{tabular}
\end{table}
\vspace{-0.8em}
\paragraph{Generalizability on Code Tasks} Beyond mathematical reasoning ability, we further evaluate our method on three code generation benchmarks: LiveCodeBench \cite{DBLP:conf/iclr/JainHGLYZWSSS25}, which is a holistic, contamination-free benchmark for evaluating code generation in LLMs, continuously updated with new problems over time; HumanEval-Pro \cite{DBLP:conf/acl/Yu0CZ25}, a variant of the widely used HumanEval \cite{DBLP:journals/corr/abs-2107-03374} suite that emphasizes self-invoking code generation task; and MBPP-Pro \cite{DBLP:conf/acl/Yu0CZ25}, an extended version of the Mostly Basic Python Problems benchmark \cite{DBLP:journals/corr/abs-2108-07732}, designed to evaluate the progressive reasoning and problem-solving capabilities of LLMs. We utilize Ling-Coder-SFT dataset\cite{codefuse2025samplemattersleveragingmixtureofexperts} as calibration and training corpus, which is a high-quality instruction-tuning corpus tailored for code generation, ensuring fair comparison across all methods. 

As shown in Table \ref{qwen2.5-7b-code}, performance on code benchmarks largely converges at checkpoint (ckpt) 20, with ckpt20 and ckpt100 yielding nearly identical results. For simplicity, we initialize streaming merging with Qwen2.5-7B and ckpt20. ARM achieves the highest average score of 48.2, surpassing TA by 0.2 points and the ckpt20 by \textbf{1.3}. The improvement demonstrates that rotation-aware merging can further enhance even saturated models in code generation tasks, validating the generalizability of our approach.

\begin{figure}[htbp]
    \centering
    \includegraphics[width=0.75\linewidth]{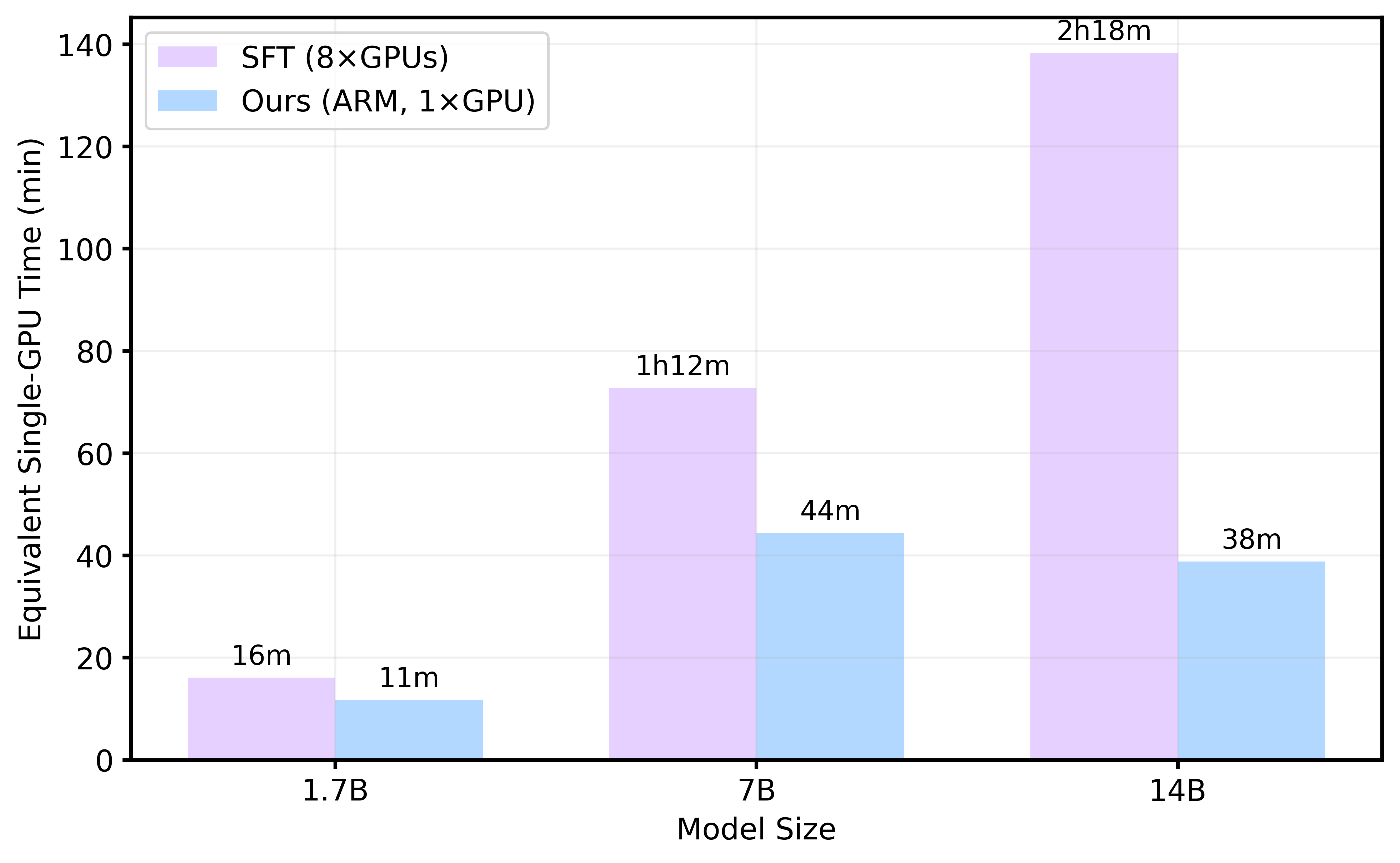}
    \caption{Time consumption analysis: SFT vs ARM}
    \label{fig:latency}
\end{figure}

\paragraph{Latency Analysis}
In Fig.~\ref{fig:latency}, we report the time costs across models of varying scales. For the 1.7B and 7B models, training had not yet converged at checkpoint 20; thus, we measure the time from this point to training convergence, as well as the subsequent merging convergence time. In contrast, the 14B model converges by checkpoint 20, yet our method continues to improve performance post-convergence; hence, we report the full training and merging durations. For 1.7B and 7B models, ARM converges in approximately \textbf{five} iterations; for the 14B model, only \textbf{two} suffice. Notably, our approach operates on a single GPU, whereas training requires 8 GPUs, and the time savings increase with model scale.

\vspace{-1em}

\section{Conclusion}
We propose Streaming Merging, a dynamic paradigm that reinterprets model merging as an iterative optimization process. By replacing the arithmetic operator with ARM, our method approximates training dynamics via activation-derived rotations. Experiments show ARM not only revitalizes early checkpoints but surpasses fully converged models on diverse tasks, demonstrating that data-driven merging can extend optimization trajectories, thus offering a scalable, training-free pathway for continuous model evolution.

\section*{Impact Statement}

This paper presents work whose goal is to advance the field of Machine
Learning. There are many potential societal consequences of our work, none
which we feel must be specifically highlighted here.






\bibliography{example_paper}

\begin{thebibliography}{41}
\providecommand{\natexlab}[1]{#1}
\providecommand{\url}[1]{\texttt{#1}}
\expandafter\ifx\csname urlstyle\endcsname\relax
  \providecommand{\doi}[1]{doi: #1}\else
  \providecommand{\doi}{doi: \begingroup \urlstyle{rm}\Url}\fi

\bibitem[Austin et~al.(2021)Austin, Odena, Nye, Bosma, Michalewski, Dohan, Jiang, Cai, Terry, Le, and Sutton]{DBLP:journals/corr/abs-2108-07732}
Austin, J., Odena, A., Nye, M.~I., Bosma, M., Michalewski, H., Dohan, D., Jiang, E., Cai, C.~J., Terry, M., Le, Q.~V., and Sutton, C.
\newblock Program synthesis with large language models.
\newblock \emph{CoRR}, abs/2108.07732, 2021.
\newblock URL \url{https://arxiv.org/abs/2108.07732}.

\bibitem[Cai et~al.(2025)Cai, Cao, Xu, Yao, Huang, Tan, Zhang, Liu, and Fang]{DBLP:journals/corr/abs-2510-00553}
Cai, Y., Cao, D., Xu, X., Yao, Z., Huang, Y., Tan, Z., Zhang, B., Liu, G., and Fang, J.
\newblock On predictability of reinforcement learning dynamics for large language models.
\newblock \emph{CoRR}, abs/2510.00553, 2025.
\newblock \doi{10.48550/ARXIV.2510.00553}.
\newblock URL \url{https://doi.org/10.48550/arXiv.2510.00553}.

\bibitem[Chen et~al.(2021)Chen, Tworek, Jun, Yuan, de~Oliveira~Pinto, Kaplan, Edwards, Burda, Joseph, Brockman, Ray, Puri, Krueger, Petrov, Khlaaf, Sastry, Mishkin, Chan, Gray, Ryder, Pavlov, Power, Kaiser, Bavarian, Winter, Tillet, Such, Cummings, Plappert, Chantzis, Barnes, Herbert{-}Voss, Guss, Nichol, Paino, Tezak, Tang, Babuschkin, Balaji, Jain, Saunders, Hesse, Carr, Leike, Achiam, Misra, Morikawa, Radford, Knight, Brundage, Murati, Mayer, Welinder, McGrew, Amodei, McCandlish, Sutskever, and Zaremba]{DBLP:journals/corr/abs-2107-03374}
Chen, M., Tworek, J., Jun, H., Yuan, Q., de~Oliveira~Pinto, H.~P., Kaplan, J., Edwards, H., Burda, Y., Joseph, N., Brockman, G., Ray, A., Puri, R., Krueger, G., Petrov, M., Khlaaf, H., Sastry, G., Mishkin, P., Chan, B., Gray, S., Ryder, N., Pavlov, M., Power, A., Kaiser, L., Bavarian, M., Winter, C., Tillet, P., Such, F.~P., Cummings, D., Plappert, M., Chantzis, F., Barnes, E., Herbert{-}Voss, A., Guss, W.~H., Nichol, A., Paino, A., Tezak, N., Tang, J., Babuschkin, I., Balaji, S., Jain, S., Saunders, W., Hesse, C., Carr, A.~N., Leike, J., Achiam, J., Misra, V., Morikawa, E., Radford, A., Knight, M., Brundage, M., Murati, M., Mayer, K., Welinder, P., McGrew, B., Amodei, D., McCandlish, S., Sutskever, I., and Zaremba, W.
\newblock Evaluating large language models trained on code.
\newblock \emph{CoRR}, abs/2107.03374, 2021.
\newblock URL \url{https://arxiv.org/abs/2107.03374}.

\bibitem[Cobbe et~al.(2021)Cobbe, Kosaraju, Bavarian, Chen, Jun, Kaiser, Plappert, Tworek, Hilton, Nakano, Hesse, and Schulman]{DBLP:journals/corr/abs-2110-14168}
Cobbe, K., Kosaraju, V., Bavarian, M., Chen, M., Jun, H., Kaiser, L., Plappert, M., Tworek, J., Hilton, J., Nakano, R., Hesse, C., and Schulman, J.
\newblock Training verifiers to solve math word problems.
\newblock \emph{CoRR}, abs/2110.14168, 2021.
\newblock URL \url{https://arxiv.org/abs/2110.14168}.

\bibitem[Codefuse \& Team(2025)Codefuse and Team]{codefuse2025samplemattersleveragingmixtureofexperts}
Codefuse and Team, L.
\newblock Every sample matters: Leveraging mixture-of-experts and high-quality data for efficient and accurate code llm, 2025.
\newblock URL \url{https://arxiv.org/abs/2503.17793}.

\bibitem[Comanici et~al.(2025)Comanici, Bieber, and et~al.]{comanici2025gemini25pushingfrontier}
Comanici, G., Bieber, E., and et~al.
\newblock Gemini 2.5: Pushing the frontier with advanced reasoning, multimodality, long context, and next generation agentic capabilities, 2025.
\newblock URL \url{https://arxiv.org/abs/2507.06261}.

\bibitem[Fei et~al.(2023)Fei, Wei, Liu, Li, and Chen]{fei2023surveygeometricoptimizationdeep}
Fei, Y., Wei, X., Liu, Y., Li, Z., and Chen, M.
\newblock A survey of geometric optimization for deep learning: From euclidean space to riemannian manifold, 2023.
\newblock URL \url{https://arxiv.org/abs/2302.08210}.

\bibitem[He et~al.(2024)He, Luo, Bai, Hu, Thai, Shen, Hu, Han, Huang, Zhang, Liu, Qi, Liu, and Sun]{DBLP:conf/acl/HeLBHTSHHHZLQL024}
He, C., Luo, R., Bai, Y., Hu, S., Thai, Z.~L., Shen, J., Hu, J., Han, X., Huang, Y., Zhang, Y., Liu, J., Qi, L., Liu, Z., and Sun, M.
\newblock Olympiadbench: {A} challenging benchmark for promoting {AGI} with olympiad-level bilingual multimodal scientific problems.
\newblock In Ku, L., Martins, A., and Srikumar, V. (eds.), \emph{Proceedings of the 62nd Annual Meeting of the Association for Computational Linguistics (Volume 1: Long Papers), {ACL} 2024, Bangkok, Thailand, August 11-16, 2024}, pp.\  3828--3850. Association for Computational Linguistics, 2024.
\newblock \doi{10.18653/V1/2024.ACL-LONG.211}.
\newblock URL \url{https://doi.org/10.18653/v1/2024.acl-long.211}.

\bibitem[Hu et~al.(2022)Hu, Shen, Wallis, Allen{-}Zhu, Li, Wang, Wang, and Chen]{DBLP:conf/iclr/HuSWALWWC22}
Hu, E.~J., Shen, Y., Wallis, P., Allen{-}Zhu, Z., Li, Y., Wang, S., Wang, L., and Chen, W.
\newblock Lora: Low-rank adaptation of large language models.
\newblock In \emph{The Tenth International Conference on Learning Representations, {ICLR} 2022, Virtual Event, April 25-29, 2022}. OpenReview.net, 2022.
\newblock URL \url{https://openreview.net/forum?id=nZeVKeeFYf9}.

\bibitem[Ilharco et~al.(2023)Ilharco, Ribeiro, Wortsman, Schmidt, Hajishirzi, and Farhadi]{DBLP:conf/iclr/IlharcoRWSHF23}
Ilharco, G., Ribeiro, M.~T., Wortsman, M., Schmidt, L., Hajishirzi, H., and Farhadi, A.
\newblock Editing models with task arithmetic.
\newblock In \emph{The Eleventh International Conference on Learning Representations, {ICLR} 2023, Kigali, Rwanda, May 1-5, 2023}. OpenReview.net, 2023.
\newblock URL \url{https://openreview.net/forum?id=6t0Kwf8-jrj}.

\bibitem[Izmailov et~al.(2018)Izmailov, Podoprikhin, Garipov, Vetrov, and Wilson]{DBLP:conf/uai/IzmailovPGVW18}
Izmailov, P., Podoprikhin, D., Garipov, T., Vetrov, D.~P., and Wilson, A.~G.
\newblock Averaging weights leads to wider optima and better generalization.
\newblock In Globerson, A. and Silva, R. (eds.), \emph{Proceedings of the Thirty-Fourth Conference on Uncertainty in Artificial Intelligence, {UAI} 2018, Monterey, California, USA, August 6-10, 2018}, pp.\  876--885. {AUAI} Press, 2018.
\newblock URL \url{http://auai.org/uai2018/proceedings/papers/313.pdf}.

\bibitem[Jain et~al.(2025)Jain, Han, Gu, Li, Yan, Zhang, Wang, Solar{-}Lezama, Sen, and Stoica]{DBLP:conf/iclr/JainHGLYZWSSS25}
Jain, N., Han, K., Gu, A., Li, W., Yan, F., Zhang, T., Wang, S., Solar{-}Lezama, A., Sen, K., and Stoica, I.
\newblock Livecodebench: Holistic and contamination free evaluation of large language models for code.
\newblock In \emph{The Thirteenth International Conference on Learning Representations, {ICLR} 2025, Singapore, April 24-28, 2025}. OpenReview.net, 2025.
\newblock URL \url{https://openreview.net/forum?id=chfJJYC3iL}.

\bibitem[Jang et~al.(2024)Jang, Huynh, Shah, Chen, and Lim]{DBLP:conf/eccv/JangHSCL24}
Jang, Y.~K., Huynh, D., Shah, A., Chen, W., and Lim, S.
\newblock Spherical linear interpolation and text-anchoring for zero-shot composed image retrieval.
\newblock In Leonardis, A., Ricci, E., Roth, S., Russakovsky, O., Sattler, T., and Varol, G. (eds.), \emph{Computer Vision - {ECCV} 2024 - 18th European Conference, Milan, Italy, September 29-October 4, 2024, Proceedings, Part {XIX}}, volume 15077 of \emph{Lecture Notes in Computer Science}, pp.\  239--254. Springer, 2024.
\newblock \doi{10.1007/978-3-031-72655-2\_14}.
\newblock URL \url{https://doi.org/10.1007/978-3-031-72655-2\_14}.

\bibitem[Kaddour(2022)]{DBLP:journals/corr/abs-2209-14981}
Kaddour, J.
\newblock Stop wasting my time! saving days of imagenet and {BERT} training with latest weight averaging.
\newblock \emph{CoRR}, abs/2209.14981, 2022.
\newblock \doi{10.48550/ARXIV.2209.14981}.
\newblock URL \url{https://doi.org/10.48550/arXiv.2209.14981}.

\bibitem[Lewkowycz et~al.(2022)Lewkowycz, Andreassen, Dohan, Dyer, Michalewski, Ramasesh, Slone, Anil, Schlag, Gutman-Solo, Wu, Neyshabur, Gur-Ari, and Misra]{lewkowycz2022solvingquantitativereasoningproblems}
Lewkowycz, A., Andreassen, A., Dohan, D., Dyer, E., Michalewski, H., Ramasesh, V., Slone, A., Anil, C., Schlag, I., Gutman-Solo, T., Wu, Y., Neyshabur, B., Gur-Ari, G., and Misra, V.
\newblock Solving quantitative reasoning problems with language models, 2022.
\newblock URL \url{https://arxiv.org/abs/2206.14858}.

\bibitem[LI et~al.(2024)LI, Beeching, Tunstall, Lipkin, Soletskyi, Huang, Rasul, Yu, Jiang, Shen, Qin, Dong, Zhou, Fleureau, Lample, and Polu]{numina_math_datasets}
LI, J., Beeching, E., Tunstall, L., Lipkin, B., Soletskyi, R., Huang, S.~C., Rasul, K., Yu, L., Jiang, A., Shen, Z., Qin, Z., Dong, B., Zhou, L., Fleureau, Y., Lample, G., and Polu, S.
\newblock Numinamath.
\newblock \url{[https://huggingface.co/AI-MO/NuminaMath-CoT](https://github.com/project-numina/aimo-progress-prize/blob/main/report/numina_dataset.pdf)}, 2024.

\bibitem[Li et~al.(2025)Li, Ma, Yan, Zhang, Liu, Lu, Xu, Chen, Wang, Zhan, Ma, Lai, Liu, Luo, Bin, Ren, Han, Hao, Yi, Liu, Ma, Jia, Zhou, Qiao, Xiang, and Wu]{DBLP:journals/corr/abs-2505-12082}
Li, Y., Ma, Y., Yan, S., Zhang, C., Liu, J., Lu, J., Xu, Z., Chen, M., Wang, M., Zhan, S., Ma, J., Lai, X., Liu, D., Luo, Y., Bin, X., Ren, H., Han, M., Hao, W., Yi, B., Liu, L., Ma, B., Jia, X., Zhou, X., Qiao, S., Xiang, L., and Wu, Y.
\newblock Model merging in pre-training of large language models.
\newblock \emph{CoRR}, abs/2505.12082, 2025.
\newblock \doi{10.48550/ARXIV.2505.12082}.
\newblock URL \url{https://doi.org/10.48550/arXiv.2505.12082}.

\bibitem[Liu et~al.(2025{\natexlab{a}})Liu, Wu, He, Han, Yuan, and Song]{DBLP:conf/acl/00010HHYS25}
Liu, S., Wu, H., He, B., Han, X., Yuan, M., and Song, L.
\newblock Sens-merging: Sensitivity-guided parameter balancing for merging large language models.
\newblock In Che, W., Nabende, J., Shutova, E., and Pilehvar, M.~T. (eds.), \emph{Findings of the Association for Computational Linguistics, {ACL} 2025, Vienna, Austria, July 27 - August 1, 2025}, pp.\  19243--19255. Association for Computational Linguistics, 2025{\natexlab{a}}.
\newblock URL \url{https://aclanthology.org/2025.findings-acl.984/}.

\bibitem[Liu et~al.(2025{\natexlab{b}})Liu, Fu, Luo, Zhu, Cheng, Hsieh, and You]{DBLP:journals/corr/abs-2510-06627}
Liu, Y., Fu, D., Luo, Y., Zhu, Z., Cheng, M., Hsieh, C., and You, Y.
\newblock {POME:} post optimization model edit via muon-style projection.
\newblock \emph{CoRR}, abs/2510.06627, 2025{\natexlab{b}}.
\newblock \doi{10.48550/ARXIV.2510.06627}.
\newblock URL \url{https://doi.org/10.48550/arXiv.2510.06627}.

\bibitem[Liu et~al.(2025{\natexlab{c}})Liu, Wu, Yao, She, Han, Zhong, and Yuan]{DBLP:journals/corr/abs-2502-10749}
Liu, Z., Wu, H., Yao, Y., She, R., Han, X., Zhong, T., and Yuan, M.
\newblock Lore-merging: Exploring low-rank estimation for large language model merging.
\newblock \emph{CoRR}, abs/2502.10749, 2025{\natexlab{c}}.
\newblock \doi{10.48550/ARXIV.2502.10749}.
\newblock URL \url{https://doi.org/10.48550/arXiv.2502.10749}.

\bibitem[Lu et~al.(2024)Lu, Fan, Wei, Qu, Chen, and Cheng]{lu2024twinmergingdynamicintegrationmodular}
Lu, Z., Fan, C., Wei, W., Qu, X., Chen, D., and Cheng, Y.
\newblock Twin-merging: Dynamic integration of modular expertise in model merging, 2024.
\newblock URL \url{https://arxiv.org/abs/2406.15479}.

\bibitem[Meng et~al.(2022)Meng, Bau, Andonian, and Belinkov]{DBLP:conf/nips/MengBAB22}
Meng, K., Bau, D., Andonian, A., and Belinkov, Y.
\newblock Locating and editing factual associations in {GPT}.
\newblock In Koyejo, S., Mohamed, S., Agarwal, A., Belgrave, D., Cho, K., and Oh, A. (eds.), \emph{Advances in Neural Information Processing Systems 35: Annual Conference on Neural Information Processing Systems 2022, NeurIPS 2022, New Orleans, LA, USA, November 28 - December 9, 2022}, 2022.

\bibitem[Morales{-}Brotons et~al.(2024)Morales{-}Brotons, Vogels, and Hendrikx]{DBLP:journals/tmlr/Morales-Brotons24}
Morales{-}Brotons, D., Vogels, T., and Hendrikx, H.
\newblock Exponential moving average of weights in deep learning: Dynamics and benefits.
\newblock \emph{Trans. Mach. Learn. Res.}, 2024, 2024.
\newblock URL \url{https://openreview.net/forum?id=2M9CUnYnBA}.

\bibitem[Neyshabur et~al.(2017)Neyshabur, Tomioka, Salakhutdinov, and Srebro]{neyshabur2017geometryoptimizationimplicitregularization}
Neyshabur, B., Tomioka, R., Salakhutdinov, R., and Srebro, N.
\newblock Geometry of optimization and implicit regularization in deep learning, 2017.
\newblock URL \url{https://arxiv.org/abs/1705.03071}.

\bibitem[Nobari et~al.(2025)Nobari, Alimohammadi, ArjomandBigdeli, Srivastava, Ahmed, and Azizan]{DBLP:journals/corr/abs-2502-02421}
Nobari, A.~H., Alimohammadi, K., ArjomandBigdeli, A., Srivastava, A., Ahmed, F., and Azizan, N.
\newblock Activation-informed merging of large language models.
\newblock \emph{CoRR}, abs/2502.02421, 2025.
\newblock \doi{10.48550/ARXIV.2502.02421}.
\newblock URL \url{https://doi.org/10.48550/arXiv.2502.02421}.

\bibitem[Pham \& Nguyen(2024)Pham and Nguyen]{DBLP:conf/emnlp/PhamN24}
Pham, V. and Nguyen, T.
\newblock Householder pseudo-rotation: {A} novel approach to activation editing in llms with direction-magnitude perspective.
\newblock In Al{-}Onaizan, Y., Bansal, M., and Chen, Y. (eds.), \emph{Proceedings of the 2024 Conference on Empirical Methods in Natural Language Processing, {EMNLP} 2024, Miami, FL, USA, November 12-16, 2024}, pp.\  13737--13751. Association for Computational Linguistics, 2024.
\newblock \doi{10.18653/V1/2024.EMNLP-MAIN.761}.
\newblock URL \url{https://doi.org/10.18653/v1/2024.emnlp-main.761}.

\bibitem[Shao et~al.(2024)Shao, Wang, Zhu, Xu, Song, Zhang, Li, Wu, and Guo]{DBLP:journals/corr/abs-2402-03300}
Shao, Z., Wang, P., Zhu, Q., Xu, R., Song, J., Zhang, M., Li, Y.~K., Wu, Y., and Guo, D.
\newblock Deepseekmath: Pushing the limits of mathematical reasoning in open language models.
\newblock \emph{CoRR}, abs/2402.03300, 2024.
\newblock \doi{10.48550/ARXIV.2402.03300}.
\newblock URL \url{https://doi.org/10.48550/arXiv.2402.03300}.

\bibitem[Shoemake(1985)]{DBLP:conf/siggraph/Shoemake85}
Shoemake, K.
\newblock Animating rotation with quaternion curves.
\newblock In Cole, P., Heilman, R., and Barsky, B.~A. (eds.), \emph{Proceedings of the 12th Annual Conference on Computer Graphics and Interactive Techniques, {SIGGRAPH} 1985, San Francisco, California, USA, July 22-26, 1985}, pp.\  245--254. {ACM}, 1985.
\newblock \doi{10.1145/325334.325242}.
\newblock URL \url{https://doi.org/10.1145/325334.325242}.

\bibitem[Tang et~al.(2025)Tang, Yang, Shen, Luo, Hu, Du, and Tao]{DBLP:journals/corr/abs-2501-09522}
Tang, A., Yang, E., Shen, L., Luo, Y., Hu, H., Du, B., and Tao, D.
\newblock Merging models on the fly without retraining: {A} sequential approach to scalable continual model merging.
\newblock \emph{CoRR}, abs/2501.09522, 2025.
\newblock \doi{10.48550/ARXIV.2501.09522}.
\newblock URL \url{https://doi.org/10.48550/arXiv.2501.09522}.

\bibitem[Tang et~al.(2024)Tang, Zhang, Wang, and Wei]{DBLP:conf/icml/TangZWW24}
Tang, Z., Zhang, X., Wang, B., and Wei, F.
\newblock Mathscale: Scaling instruction tuning for mathematical reasoning.
\newblock In \emph{Forty-first International Conference on Machine Learning, {ICML} 2024, Vienna, Austria, July 21-27, 2024}. OpenReview.net, 2024.
\newblock URL \url{https://openreview.net/forum?id=Kjww7ZN47M}.

\bibitem[Vershynin(2018)]{vershynin2018high}
Vershynin, R.
\newblock \emph{High-dimensional probability: An introduction with applications in data science}, volume~47.
\newblock Cambridge university press, 2018.

\bibitem[Wortsman et~al.(2022)Wortsman, Ilharco, Gadre, Roelofs, Lopes, Morcos, Namkoong, Farhadi, Carmon, Kornblith, and Schmidt]{DBLP:conf/icml/WortsmanIGRLMNF22}
Wortsman, M., Ilharco, G., Gadre, S.~Y., Roelofs, R., Lopes, R.~G., Morcos, A.~S., Namkoong, H., Farhadi, A., Carmon, Y., Kornblith, S., and Schmidt, L.
\newblock Model soups: averaging weights of multiple fine-tuned models improves accuracy without increasing inference time.
\newblock In Chaudhuri, K., Jegelka, S., Song, L., Szepesv{\'{a}}ri, C., Niu, G., and Sabato, S. (eds.), \emph{International Conference on Machine Learning, {ICML} 2022, 17-23 July 2022, Baltimore, Maryland, {USA}}, volume 162 of \emph{Proceedings of Machine Learning Research}, pp.\  23965--23998. {PMLR}, 2022.
\newblock URL \url{https://proceedings.mlr.press/v162/wortsman22a.html}.

\bibitem[Wu et~al.(2025)Wu, Yang, Liu, Wang, Xu, and Wong]{DBLP:journals/corr/abs-2509-23595}
Wu, T., Yang, R., Liu, T., Wang, J., Xu, Z., and Wong, N.
\newblock Timber: Training-free instruct model refining with base via effective rank.
\newblock \emph{CoRR}, abs/2509.23595, 2025.
\newblock \doi{10.48550/ARXIV.2509.23595}.
\newblock URL \url{https://doi.org/10.48550/arXiv.2509.23595}.

\bibitem[Yadav et~al.(2023)Yadav, Tam, Choshen, Raffel, and Bansal]{DBLP:conf/nips/YadavTCRB23}
Yadav, P., Tam, D., Choshen, L., Raffel, C.~A., and Bansal, M.
\newblock Ties-merging: Resolving interference when merging models.
\newblock In Oh, A., Naumann, T., Globerson, A., Saenko, K., Hardt, M., and Levine, S. (eds.), \emph{Advances in Neural Information Processing Systems 36: Annual Conference on Neural Information Processing Systems 2023, NeurIPS 2023, New Orleans, LA, USA, December 10 - 16, 2023}, 2023.

\bibitem[Yang et~al.(2024{\natexlab{a}})Yang, Yang, Zhang, Hui, Zheng, Yu, Li, Liu, Huang, Wei, Lin, Yang, Tu, Zhang, Yang, Yang, Zhou, Lin, Dang, Lu, Bao, Yang, Yu, Li, Xue, Zhang, Zhu, Men, Lin, Li, Xia, Ren, Ren, Fan, Su, Zhang, Wan, Liu, Cui, Zhang, and Qiu]{DBLP:journals/corr/abs-2412-15115}
Yang, A., Yang, B., Zhang, B., Hui, B., Zheng, B., Yu, B., Li, C., Liu, D., Huang, F., Wei, H., Lin, H., Yang, J., Tu, J., Zhang, J., Yang, J., Yang, J., Zhou, J., Lin, J., Dang, K., Lu, K., Bao, K., Yang, K., Yu, L., Li, M., Xue, M., Zhang, P., Zhu, Q., Men, R., Lin, R., Li, T., Xia, T., Ren, X., Ren, X., Fan, Y., Su, Y., Zhang, Y., Wan, Y., Liu, Y., Cui, Z., Zhang, Z., and Qiu, Z.
\newblock Qwen2.5 technical report.
\newblock \emph{CoRR}, abs/2412.15115, 2024{\natexlab{a}}.
\newblock \doi{10.48550/ARXIV.2412.15115}.
\newblock URL \url{https://doi.org/10.48550/arXiv.2412.15115}.

\bibitem[Yang et~al.(2025{\natexlab{a}})Yang, Li, Yang, Zhang, Hui, Zheng, Yu, Gao, Huang, Lv, Zheng, Liu, Zhou, Huang, Hu, Ge, Wei, Lin, Tang, Yang, Tu, Zhang, Yang, Yang, Zhou, Lin, Dang, Bao, Yang, Yu, Deng, Li, Xue, Li, Zhang, Wang, Zhu, Men, Gao, Liu, Luo, Li, Tang, Yin, Ren, Wang, Zhang, Ren, Fan, Su, Zhang, Zhang, Wan, Liu, Wang, Cui, Zhang, Zhou, and Qiu]{DBLP:journals/corr/abs-2505-09388}
Yang, A., Li, A., Yang, B., Zhang, B., Hui, B., Zheng, B., Yu, B., Gao, C., Huang, C., Lv, C., Zheng, C., Liu, D., Zhou, F., Huang, F., Hu, F., Ge, H., Wei, H., Lin, H., Tang, J., Yang, J., Tu, J., Zhang, J., Yang, J., Yang, J., Zhou, J., Lin, J., Dang, K., Bao, K., Yang, K., Yu, L., Deng, L., Li, M., Xue, M., Li, M., Zhang, P., Wang, P., Zhu, Q., Men, R., Gao, R., Liu, S., Luo, S., Li, T., Tang, T., Yin, W., Ren, X., Wang, X., Zhang, X., Ren, X., Fan, Y., Su, Y., Zhang, Y., Zhang, Y., Wan, Y., Liu, Y., Wang, Z., Cui, Z., Zhang, Z., Zhou, Z., and Qiu, Z.
\newblock Qwen3 technical report.
\newblock \emph{CoRR}, abs/2505.09388, 2025{\natexlab{a}}.
\newblock \doi{10.48550/ARXIV.2505.09388}.
\newblock URL \url{https://doi.org/10.48550/arXiv.2505.09388}.

\bibitem[Yang et~al.(2024{\natexlab{b}})Yang, Shen, Guo, Wang, Cao, Zhang, and Tao]{DBLP:journals/corr/abs-2408-07666}
Yang, E., Shen, L., Guo, G., Wang, X., Cao, X., Zhang, J., and Tao, D.
\newblock Model merging in llms, mllms, and beyond: Methods, theories, applications and opportunities.
\newblock \emph{CoRR}, abs/2408.07666, 2024{\natexlab{b}}.
\newblock \doi{10.48550/ARXIV.2408.07666}.
\newblock URL \url{https://doi.org/10.48550/arXiv.2408.07666}.

\bibitem[Yang et~al.(2025{\natexlab{b}})Yang, Tang, Shen, Guo, Wang, Cao, and Zhang]{yang2025continual}
Yang, E., Tang, A., Shen, L., Guo, G., Wang, X., Cao, X., and Zhang, J.
\newblock Continual model merging without data: Dual projections for balancing stability and plasticity.
\newblock In \emph{The Thirty-ninth Annual Conference on Neural Information Processing Systems}, 2025{\natexlab{b}}.

\bibitem[Yao et~al.(2025)Yao, Liu, Liu, Li, Liu, Han, Guo, Wu, and Song]{DBLP:journals/corr/abs-2505-14009}
Yao, Y., Liu, S., Liu, Z., Li, Q., Liu, M., Han, X., Guo, Z., Wu, H., and Song, L.
\newblock Activation-guided consensus merging for large language models.
\newblock \emph{CoRR}, abs/2505.14009, 2025.
\newblock \doi{10.48550/ARXIV.2505.14009}.
\newblock URL \url{https://doi.org/10.48550/arXiv.2505.14009}.

\bibitem[Yu et~al.(2024)Yu, Yu, Yu, Huang, and Li]{DBLP:conf/icml/Yu0Y0L24}
Yu, L., Yu, B., Yu, H., Huang, F., and Li, Y.
\newblock Language models are super mario: Absorbing abilities from homologous models as a free lunch.
\newblock In \emph{Forty-first International Conference on Machine Learning, {ICML} 2024, Vienna, Austria, July 21-27, 2024}. OpenReview.net, 2024.
\newblock URL \url{https://openreview.net/forum?id=fq0NaiU8Ex}.

\bibitem[Yu et~al.(2025)Yu, Zhao, Cohan, and Zhang]{DBLP:conf/acl/Yu0CZ25}
Yu, Z., Zhao, Y., Cohan, A., and Zhang, X.
\newblock Humaneval pro and {MBPP} pro: Evaluating large language models on self-invoking code generation task.
\newblock In Che, W., Nabende, J., Shutova, E., and Pilehvar, M.~T. (eds.), \emph{Findings of the Association for Computational Linguistics, {ACL} 2025, Vienna, Austria, July 27 - August 1, 2025}, pp.\  13253--13279. Association for Computational Linguistics, 2025.
\newblock URL \url{https://aclanthology.org/2025.findings-acl.686/}.

\end{thebibliography}
\bibliographystyle{icml2026}

\newpage
\appendix
\onecolumn
\section{Appendix} \label{appendix}
\subsection{The upper bound of Arithmetic Streaming Merging.} \label{upper bound}

\paragraph{Sliding Window Scheme} Given the objective:
\[
W_{t} = W_{t-n} + \lambda/n \sum_{i=t-n}^{t-1}  (W_{i} - W_{t-n}), \quad t > n,
\label{eq:linear_recurrence}
\]

Specifically, suppose window size is 2, let $W_0$ and $W_1$ denote two fine-tuned checkpoints. Merging proceeds sequentially: $W_0$ and $W_1$ are merged to produce $W_2$, $W_1$ and $W_2$ yield $W_3$, $W_2$ and $W_3$ yield $W_4$, and so on. Formally, for $t \geq 2$, the update follows $W_t = \mathcal{M}(W_{t-2}, W_{t-1})$, where $\mathcal{M}$ denotes the merging operator. We start merging from these checkpoints. Fix a general scalar parameter $\lambda \in (0,1]$. The arithmetic recurrence admits the closed-form solution:
\[
W_t = \frac{(1 - \lambda) W_0 + W_1}{2 - \lambda} + \frac{W_0 - W_1}{2 - \lambda} (\lambda - 1)^n.
\]
Consequently, the sequence converges to the predictable limit, which constrains the potential of arithmetic merging:
\[
\lim_{t \to \infty} W_t = \frac{(1 - \lambda) W_0 + W_1}{2 - \lambda}.
\]

\paragraph{Anchoring Scheme} Given that:
\[
    W_t = W_{t-1} + \lambda/k \sum_{i=1}^{k} (W_i - W_{t-1}), \quad t > k,
\]
When the merging step size $\lambda$ is fixed with $0 < \lambda < 1$, the update rule can be expanded as follows:
\[
W_t = (1 - \lambda) W_{t-1} + \lambda \bar{W}_k, \quad \text{where } \bar{W}_k = \frac{1}{k} \sum_{i=1}^k W_i,
\]
induces exponential convergence of $W_t$ to the arithmetic mean $\bar{W}_k$:
\[
\lim_{t \to \infty} W_t = \bar{W}_k.
\]

Even when $\lambda$ is allowed to vary per step ($\lambda_t$), the recursion
\[
W_t = (1 - \lambda_t) W_{t-1} + \lambda_t \bar{W}_k
\]
yields a closed-form solution: 
\[
W_t = \alpha_t W_k + (1 - \alpha_t) \bar{W}_k, \quad \text{with } \alpha_t = \prod_{s=k+1}^{t} (1 - \lambda_s).
\]
Provided that $\sum_{t} \lambda_t = \infty$ (ensuring $\alpha_t \to 0$), we again have
\[
\lim_{t \to \infty} W_t = \bar{W}_k.
\]

Thus, under the arithmetic streaming merging formulation—where only the initial $k$ checkpoints are used and the merging target remains fixed at $\bar{W}_k$—the performance ceiling is fundamentally bounded by the quality of $\bar{W}_k$. Neither a constant nor an adaptive $\lambda_t$ can surpass this limit without modifying the merging objective or incorporating additional model information.

\subsection{Experimental Hyperparameters} \label{hyperparameters}
For SFT and RL, we employ the OpenRLHF framework\footnote{\url{https://github.com/OpenRLHF/OpenRLHF}} with the following hyperparameters: learning rate $=1\times10^{-5}$, batch size $=128$, number of epochs $=5$, maximum sequence length $=8192$, \texttt{bf16} precision, and DeepSpeed ZeRO-3 for distributed training. For LoRA, we use a rank of 16. In TA and other scenarios requiring weighting coefficients, we adopt a recommended scaling hyperparameter $\lambda = 0.7$. For TIES and DARE, we set the top-$k$ sparsity ratio and dropout rate to 0.2 and 0.9, respectively. For AIM, we set the relaxation parameter $\omega$ to 0.4 as suggested. For all activation-based methods, we use a batch size of 64 per step. We merge a base model and 2 candidate models for all approaches. ARM transitions into sliding window mode after 3 rounds of anchoring. The maximum number of total iterations is six rounds. For code tasks, We utilize a randomly sampled subset of 5K examples from the Ling-Coder-SFT dataset as training and calibration data.

\subsection{Performance Evolving Trajectory} \label{envolve}
\begin{figure}[htbp]
    \centering
    \includegraphics[width=0.6\linewidth]{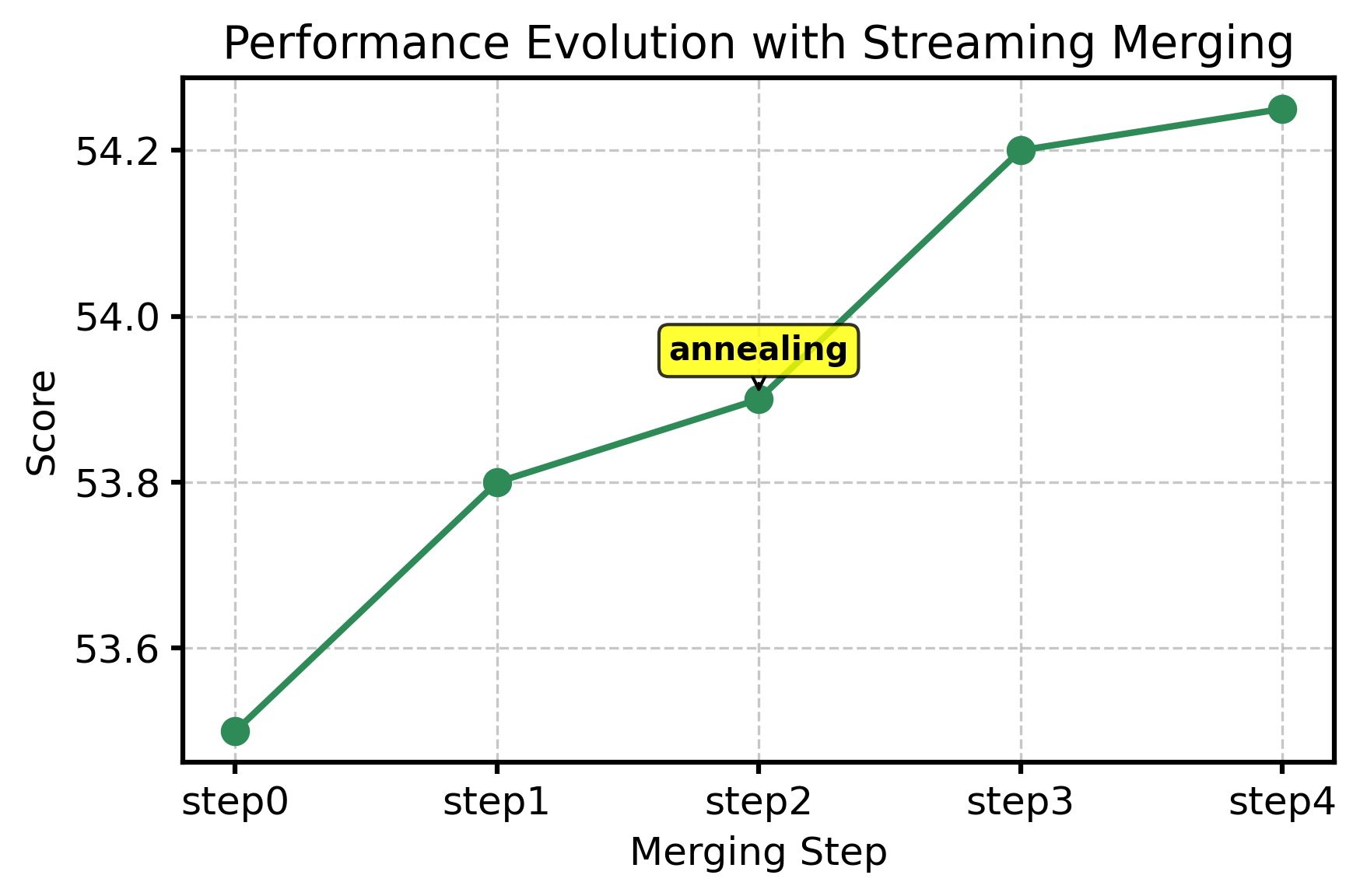}
    \caption{ARM average performance evolution on Qwen2.5-7B on the mathematical reasoning task.}
    \label{arm_envolve}
\end{figure}

As shown in Fig. \ref{arm_envolve}, starting from early-stage trained SFT checkpoints of Qwen2.5-7B, ARM embedded within Streaming Merging is inherently iterative and extensible: each merged checkpoint serves as the foundation for subsequent refinement, enabling continuous performance improvement without retraining. This property allows the process to be halted at any point based on resource constraints or convergence criteria, yet resumed later if needed—making it a flexible and sustainable paradigm for ongoing model evolution.

\subsection{Experiments on LLaMA} \label{llama}
\begin{table}
\fontsize{10}{11} \selectfont
\centering
\caption{Performance of streaming merging on LLaMA3.2-3B-Instruct during early SFT stage}
\label{llama-early}
\begin{tabular}{lcccc}
\toprule[0.8pt]
Methods &GSM8K & MATH500 & Olympiad & AVG  \\ \hline
LLaMA3.2-3B-Instruct &71.04 &40.2 &13.2 &41.48   \\
SFT ckpt 10  &65.58 &40.0 &10.09 &38.56   \\ 
SFT ckpt 20  &63.08 &39.4 &11.13 & 37.87  \\
\cdashline{1-5}
SFT ckpt 100  &71.42 &44.2 &11.42 &42.35   \\
\cdashline{1-5}
ARM &\textbf{72.48} &\textbf{46.8} &\textbf{13.95} &\textbf{44.41} \\ \bottomrule[0.8pt]
\end{tabular}
\end{table}
\begin{table}
\fontsize{10}{11} \selectfont
\centering
\caption{Performance of streaming merging on LLaMA3.2-3B-Instruct after SFT convergence}
\label{llama-converge}
\begin{tabular}{lcccc}
\toprule[0.8pt]
Methods &GSM8K & MATH500 & Olympiad & AVG  \\ \hline
LLaMA3.2-3B-Instruct &71.04 &40.2 &13.2 &41.48   \\
SFT ckpt 90  &70.66 &41.0 &10.68 &40.78   \\ 
SFT ckpt 100  &71.42 &44.2 &11.42 &42.35     \\
\cdashline{1-5}
ARM &\textbf{73.69} &\textbf{50.0 } &\textbf{15.58} &\textbf{46.42} \\ \bottomrule[0.8pt]
\end{tabular}
\end{table}

We use EvalScope\footnote{https://github.com/modelscope/evalscope} for LLaMA evaluation. During supervised fine-tuning of LLaMA3.2-3B-Instruct, the training loss exhibits high volatility early on and plateaus around checkpoint 80. Consistently, individual checkpoints at early stages (e.g., ckpt10, ckpt20) underperform the base model on GSM8K, MATH500 and OlympiadBench, indicating unstable optimization. Nevertheless, streaming merging via ARM—applied to these same early checkpoints—yields a substantially higher average score (44.41 vs. 41.48), demonstrating its ability to extract complementary signals even from immature models.

After convergence (ckpt90–100), while single checkpoints recover positive performance, ARM further improves upon them, achieving 50.0 on MATH500 and 15.58 on Olympiad, outperforming the best individual checkpoint by (4.07). This shows that merging not only rescues under-trained models but also unlocks additional gains beyond the training endpoint, offering a gradient-free pathway for continual refinement.

\subsection{Extended Experiments via LoRA}
\begin{table*}[ht!]
\fontsize{10}{11} \selectfont
\centering
\caption{Extended experiments via LoRA}
\label{qwen2.5_7b_lora}
\begin{tabular}{lcccccc}
\toprule[0.8pt]
\diagbox{Method}{Bench} & GSM8K & MATH500  & Olympiad Bench & College Math  & Minerva Math & Avg. \\
\hline
Qwen2.5-7B  &80.9 &58.6 &25.4 &38.6 &18.3 &44.4 \\ 
LoRA-ckpt10 &81.4 &58.4 &26.1 &38.5 &17.3 &44.3 \\ 
LoRA-ckpt20 &76.3 &58.4 &25.3 &35.1 &18.4 &42.7 \\ \hline
TA &81.0 &61.0 &28.5 &38.3 &17.3 &45.1\\
ARM &82.9 &58.8 &28.6 &38.3 &19.9 &45.7 \\
\bottomrule[0.8pt]
\end{tabular}
\end{table*}
In LoRA training, accuracy exhibits a non-monotonic trajectory—initially decreasing before recovering to convergence at 52.0. For simplicity and consistency with the main experiments, we evaluate on checkpoints 10 and 20 (where performance is still suboptimal), using three streaming merging iterations. Under this setting, ARM outperforms Task Arithmetic by +0.6 points on average across benchmarks, confirming its efficacy even in early, unstable training phases.

\subsection{Extended Experiments via Reinforcement Learning} \label{7b_rl}
\begin{table}
\fontsize{10}{11} \selectfont
\centering
\caption{Performance of streaming merging on Qwen2.5-7B after RL training via GRPO algorithm}
\label{qwen2.5-7b-rl}
\begin{tabular}{lcccc}
\toprule[0.8pt]
Qwen2.5-7B & MATH500 & Olympiad & Minerva & AVG  \\ \hline
RL-ckpt 40  &73.9  & 35.4  &39.6   & 50.6 \\
RL-ckpt 50  &73.6 &36.3 &40.8 &50.2       \\ \cdashline{1-5}
TA  &75.7 &\textbf{36.8} &38.6 &50.4\\
ARM (Ours) &\textbf{76.6} &36.0 &\textbf{40.4} &\textbf{51.0} \\ \bottomrule[0.8pt]
\end{tabular}
\end{table}

\paragraph{Dynamic iteration based on GRPO} As shown in Table \ref{qwen2.5-7b-rl}, on Qwen2.5-7B after GRPO-based RL training, our ARM method consistently outperforms both individual checkpoints (RL-ckpt40/50) and TA across all benchmarks, achieving an average score of 51.0, surpassing the best checkpoint by +0.4 points and TA by +0.6. This demonstrates that even in RL settings, structured streaming merging can effectively consolidate knowledge from late-stage policies to yield superior performance.

\paragraph{Combination with RL editing Approaches}
\begin{table}
\fontsize{10}{11} \selectfont
\centering
\caption{Performance of ARM combined with RL editing methods on Qwen2.5-7B}
\label{qwen2.5-7b-rl-editing}
\begin{tabular}{lcccc}
\toprule[0.8pt]
Qwen3-1.7B & MATH500 & Olympiad & Minerva & AVG  \\ \hline
Timber &75.4 &37.9 &39.3 &50.8  \\
$+$ ARM &77.8 &38.9 &38.6 &51.8  \\ \cdashline{1-5}
POME   &76.6 &33.3 &41.2 &50.4     \\
$+$ ARM &76.0 &36.3 &42.5 &51.6\\
\bottomrule[0.8pt]
\end{tabular}
\end{table}
Table \ref{qwen2.5-7b-rl-editing} demonstrates that integrating ARM with existing RL editing methods, such as Timber and POME, consistently improves performance across most benchmarks on Qwen2.5-7B. For both methods, adding ARM yields gains of +1.0$\sim$2.4 points in individual tasks and +0.8$\sim$1.2 in average score, confirming that ARM’s rotation-aware alignment enhances low-rank edits without requiring additional training.

\subsection{Ablation Study on anchoring and sliding window scheme} \label{abl_sl_achoring}
\begin{table*}[ht!]
\fontsize{9}{10} \selectfont
\centering
\caption{Ablation Study on anchoring and sliding window scheme}
\label{tab:qwen_sliding}
\begin{tabular}{lcccccc}
\toprule[0.8pt]
\diagbox{Method}{Bench} & GSM8K & MATH500  & Olympiad Bench & College Math  & Minerva Math & Avg. \\
\hline
\rowcolor{lightblue}
\multicolumn{7}{c}{\textit{Qwen3-1.7B-Base}} \\
Qwen3-1.7B-Base  &23.0 &15.0 &6.1 &7.6 &4.4 &11.2 \\ \hline
\rowcolor{lightgray}
\multicolumn{7}{c}{\textit{Training-based Methods}} \\
Checkpoint10 &58.4 &44.0 &18.5 &24.5 &10.3 &31.1 \\ 
Checkpoint20 &65.0 &51.4 &15.9 &33.9 &19.1 &37.1 \\ \cdashline{1-7}
Checkpoint150 &76.0	&55.2 &19.4	&38.1 &20.2	&41.8\\
\hline
\rowcolor{lightgray}
\multicolumn{7}{c}{\textit{Merging-based Methods}} \\
ARM(SW) &67.9 &50.0 &20.9 &30.0 &15.8 &36.9 \\ 
ARM(Achoring)  &74.5 &55.5 &23.1 &34.0 &17.3 &40.9  \\
ARM(SW+Anchoring)  &76.2 &55.2 &23.6 &33.8 &23.9 &42.5  \\
\hline
\rowcolor{lightblue}
\multicolumn{7}{c}{\textit{Qwen2.5-7B}} \\
Qwen2.5-7B  &80.9 &58.6 &25.4 &38.6 &18.3 &44.4 \\ \hline
\rowcolor{lightgray}
\multicolumn{7}{c}{\textit{Training-based Methods}} \\
Checkpoint10 &82.2 &66.3 &29.9 &41.3 &26.8 &49.3 \\ 
Checkpoint20 &84.9 &64.4 &27.1 &44.7 &29.4 &50.1 \\ \cdashline{1-7}
Checkpoint100 &86.3 &65.3 &27.0 &44.0 &33.2 &51.2 \\
\hline
\rowcolor{lightgray}
\multicolumn{7}{c}{\textit{Merging-based Methods}} \\
ARM(SW) &87.1 &68.0 &33.6 &46.5 &27.9 &52.6 \\ 
ARM(Achoring) &86.6 &71.6 &32.9 &44.8 &33.5 &53.9\\  
ARM(SW+Anchoring) &87.9 &69.6 &31.8 &46.0 &35.7 &54.2 \\
\bottomrule[0.8pt]
\end{tabular}
\end{table*}
As shown in Table~\ref{tab:qwen_sliding}, pure sliding-window merging (ARM(SW)) yields suboptimal performance, particularly on smaller models, indicating insufficient stabilization in early stages. In contrast, we observe that anchoring alone (ARM(Anchoring)) achieves strong initial gains, but plateaus prematurely, suggesting over-reliance on the initial SFT checkpoint induces representational rigidity. The hybrid scheme ARM(SW+Anchoring) consistently outperforms both variants, attaining the highest scores across benchmarks. This confirms that anchoring provides necessary directional stability, while sliding-window refinement mitigates stagnation and prevents overfitting to the initialization subspace. However, in the 14B setting, SFT converges very early, resulting in minimal geometric divergence among checkpoints. Consequently, the sliding window strategy yields no discernible improvement, and anchoring alone suffices. This further suggests that when the geometric discrepancy between merging checkpoints is small, annealing like sliding window becomes unnecessary.

\subsection{Angular and magnitude evolution of ARM} \label{rotat_mag}
\begin{figure}[htbp]
    \centering
    \includegraphics[width=\linewidth]{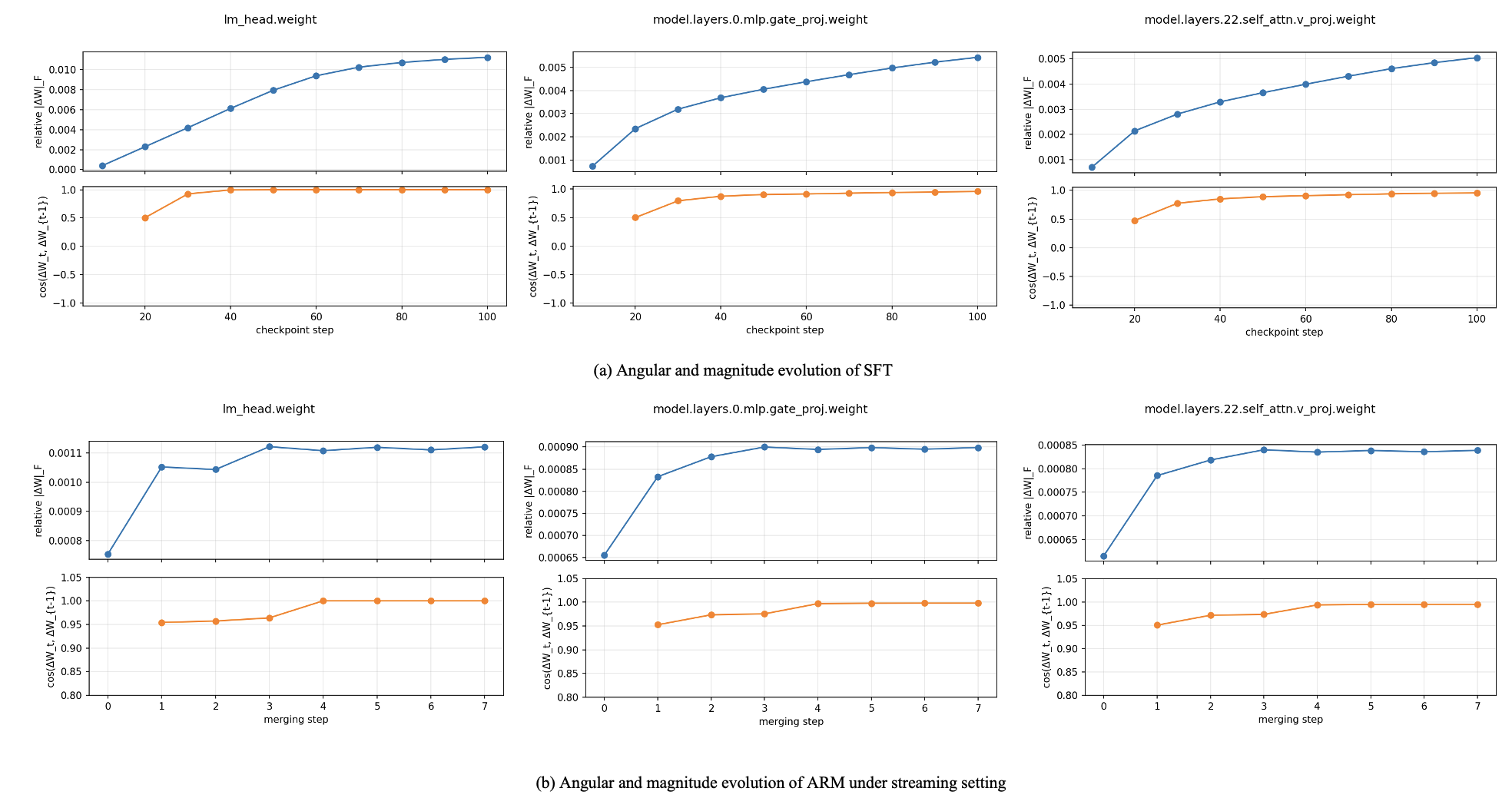}
    \caption{Angular and magnitude evolution of ARM, we randomly sample three layers from the 7B model for analysis.}
    \label{rotation_iter}
\end{figure}

As shown in the Fig. \ref{rotation_iter}, ARM exhibits strong geometric consistency under streaming merging: the update directions ($\Delta W$) maintain high cosine similarity across layers and checkpoints, indicating minimal directional drift during merging and faithfully preserving the smooth parameter evolution characteristic of SFT. Moreover, through rotation alignment, ARM approximates full fine-tuning trajectories using only a few checkpoints, highlighting its potential to substitute or extend training under resource constraints.

\subsection{ARM in Single-turn Merging Regime} \label{single ARM}
\begin{table}
\fontsize{10}{11} \selectfont
\centering
\caption{Performance of ARM on Qwen2.5-7B in single-turn merging scenario}
\label{qwen2.5-7b-multitask}
\begin{tabular}{lccc}
\toprule[0.8pt]
 &Minerva &MBPP-Pro & AVG  \\ \hline
Qwen2.5-7B  &18.3 &59.2 &38.8\\
Code-ckpt  &30.1 &59.7 &44.9  \\
Math-ckpt &32.0 &57.4 &44.7      \\ \cdashline{1-4}
TA &29.4 &58.6 &44.1\\
ARM &31.5 &58.6 &45.1
 \\ \bottomrule[0.8pt]
\end{tabular}
\end{table}
In the single-turn merging setting, we merge the Qwen2.5-7B model with two specialized SFT checkpoints—one trained on mathematical reasoning and the other on code generation, and observe that ARM consistently matches or exceeds the performance of Task Arithmetic. As shown in Table \ref{qwen2.5-7b-multitask}, ARM achieves an average score of 45.1, improving upon TA by 1.0 points. This demonstrates that ARM’s angle-aware alignment yields superior quality even in a one-shot, non-iterative regime.

\subsection{Initialization Scenario} \label{apx:init}
\begin{table*}[ht!]
\fontsize{10}{11} \selectfont
\centering
\caption{Analysis of initialization scenario}
\label{qwen2.5_7b_initial}
\begin{tabular}{lcccccc}
\toprule[0.8pt]
\diagbox{Method}{Bench} & GSM8K & MATH500  & Olympiad Bench & College Math  & Minerva Math & Avg. \\
\hline
Qwen2.5-7B  &80.9 &58.6 &25.4 &38.6 &18.3 &44.4 \\ 
Qwen2.5-7B-Instruct &91.7 &71.6 &34.1 &45.5 &37.1 &56.4 \\ \hline
Step 0 &91.5 &75.4 &32.1 &45.7 &37.1 &56.4\\
Step 1 &91.8 &74.0 &34.7 &45.1 &38.2 &56.8\\
Step 2 &91.5 &73.0 &34.4 &45.7 &39.7 &56.9\\
\bottomrule[0.8pt]
\end{tabular}
\end{table*}

Beyond using a trained initial checkpoint, we explore alternative low-cost initialization strategies to assess the robustness and generalizability of Streaming Merging across diverse starting points. We first prune the base model and merge it with the original; however, as pruning aims at compression rather than performance enhancement, the resulting structural mismatch yields negligible gains.  

To further broaden the initialization setting, we apply Streaming Merging to Qwen2.5-7B and its instruction-tuned variant, Qwen2.5-7B-Instruct. As shown in Table \ref{qwen2.5_7b_initial}, while both models perform comparably on mathematical reasoning benchmarks, two additional merging steps steadily improve performance to 56.9—surpassing the standalone Instruct model (56.4). This demonstrates that Streaming Merging remains effective even when initialized from off-trajectory models.





\end{document}